\documentclass[a4paper]{article}

\usepackage{graphicx}
\usepackage{multirow}
\usepackage{amsmath,amssymb,amsfonts}
\usepackage{amsthm}
\usepackage{hyperref}
\usepackage{mathrsfs}
\usepackage[title]{appendix}
\usepackage{xcolor}
\usepackage{textcomp}
\usepackage{manyfoot}
\usepackage{booktabs}
\usepackage{listings}

\usepackage{amssymb}
\usepackage{amsmath}
\usepackage{graphicx}
\usepackage{multirow}
\usepackage{adjustbox}
\usepackage[caption=false]{subfig}
\usepackage{url}
\usepackage{placeins}

\usepackage{makecell}

\usepackage{algorithm}
\usepackage{algpseudocode}

\usepackage{anyfontsize}

\usepackage[numbers,sort&compress]{natbib}

\graphicspath{{./assets/}}

\title{ProtoTSNet: Interpretable Multivariate Time Series Classification With Prototypical Parts}

\author{Bartłomiej Małkus$^1$ \and Szymon Bobek$^2$ \and Grzegorz J. Nalepa$^2$}

\date{
	$^1$ Doctoral School of Exact and Natural Sciences \\ Jagiellonian University \\
    Cracow, Poland \\ \texttt{bartlomiej.malkus@doctoral.uj.edu.pl} \\%
	$^2$ Faculty of Physics, Astronomy and Applied Computer Science, Institute of Applied Computer Science, and Jagiellonian Human-Centered AI Lab \\ Jagiellonian University \\
    Cracow, Poland \\ \texttt{\{szymon.bobek, grzegorz.j.nalepa\}@uj.edu.pl}\\[2ex]%
}

\begin{document}

\maketitle

\begin{abstract}
Time series data is one of the most popular data modalities in critical domains such as industry and medicine. The demand for algorithms that not only exhibit high accuracy but also offer interpretability is crucial in such fields, as decisions made there bear significant consequences. In this paper, we present ProtoTSNet, a novel approach to interpretable classification of multivariate time series data, through substantial enhancements to the ProtoPNet architecture. Our~method is tailored to overcome the unique challenges of time series analysis, including capturing dynamic patterns and handling varying feature significance. Central~to our innovation is a modified convolutional encoder utilizing group convolutions, pre-trainable as part of an autoencoder and designed to preserve and quantify feature importance. We evaluated our model on 30 multivariate time series datasets from the UEA archive, comparing our approach with existing explainable methods as well as non-explainable baselines. Through comprehensive evaluation and ablation studies, we demonstrate that our approach achieves the best performance among ante-hoc explainable methods while maintaining competitive performance with non-explainable and post-hoc explainable approaches, providing interpretable results accessible to domain experts.
\end{abstract}

\section{Introduction}

In recent years, the field of \textit{time series classification} (TSC) has received significant attention in diverse domains, including healthcare~\citep{kaushik2020ai}, finance~\citep{sezer2020financial}, and industry~\citep{keleko2022artificial}. While recent research in this field has focused primarily on improving model accuracy, aspects such as explainability and computational efficiency have often been overlooked. Although state-of-the-art models achieve impressive performance on benchmark datasets, they remain largely opaque and frequently face scalability challenges~\citep{middlehurst2023bake,cabello2024fast}.

This opacity stands in stark contrast to emerging requirements for AI transparency and explainability, initially outlined by~\cite{darpa} and subsequently mandated through regulatory frameworks such as GDPR~\citep{goodman2016regulations} and the more recent AI Act~\citep{aiact2022hacker}. Our work addresses this gap by developing an explainable TSC approach while maintaining competitive performance and scalability.

Explainable methods in machine learning are commonly categorized into post hoc and ante hoc approaches. Post hoc methods provide retrospective explanations of a model's decisions while the model itself remains a black box, an approach particularly common in neural networks~\citep{mochaourab2022post}. However, these explanations may be inaccurate or lead to misconceptions about the model's operations~\citep{rashomon2023ecml}. In contrast, ante hoc architectures incorporate explainability directly into the model's design, offering a more transparent and immediate understanding of the decision-making processes. Among these, prototype-based methods have gained significant attention, with ProtoPNet~\citep{chen2019looks} being a notable example. These~methods explain decisions through a set of learned prototypes, providing interpretable insights by highlighting key representative patterns that influence predictions.

Following recent surveys in the field of explainable time series classification~\citep{theissler2022explainable}, two significant observations can be made. First, a large part of existing studies focus on univariate time series. 
Multivariate data presents additional complexities, including the potential inter-feature dependencies and feature redundancy. 
Secondly, a large part of the studies do not provide a way to reproduce the results, which is imposing what is becoming known as the replication crisis~\citep{kapoor2023replication}.
In the context of explainability, we strive for transparency, and the ability to replicate findings is a crucial component of transparent research.

\begin{figure}[htbp]
    \centering
    \includegraphics[width=0.85\textwidth]{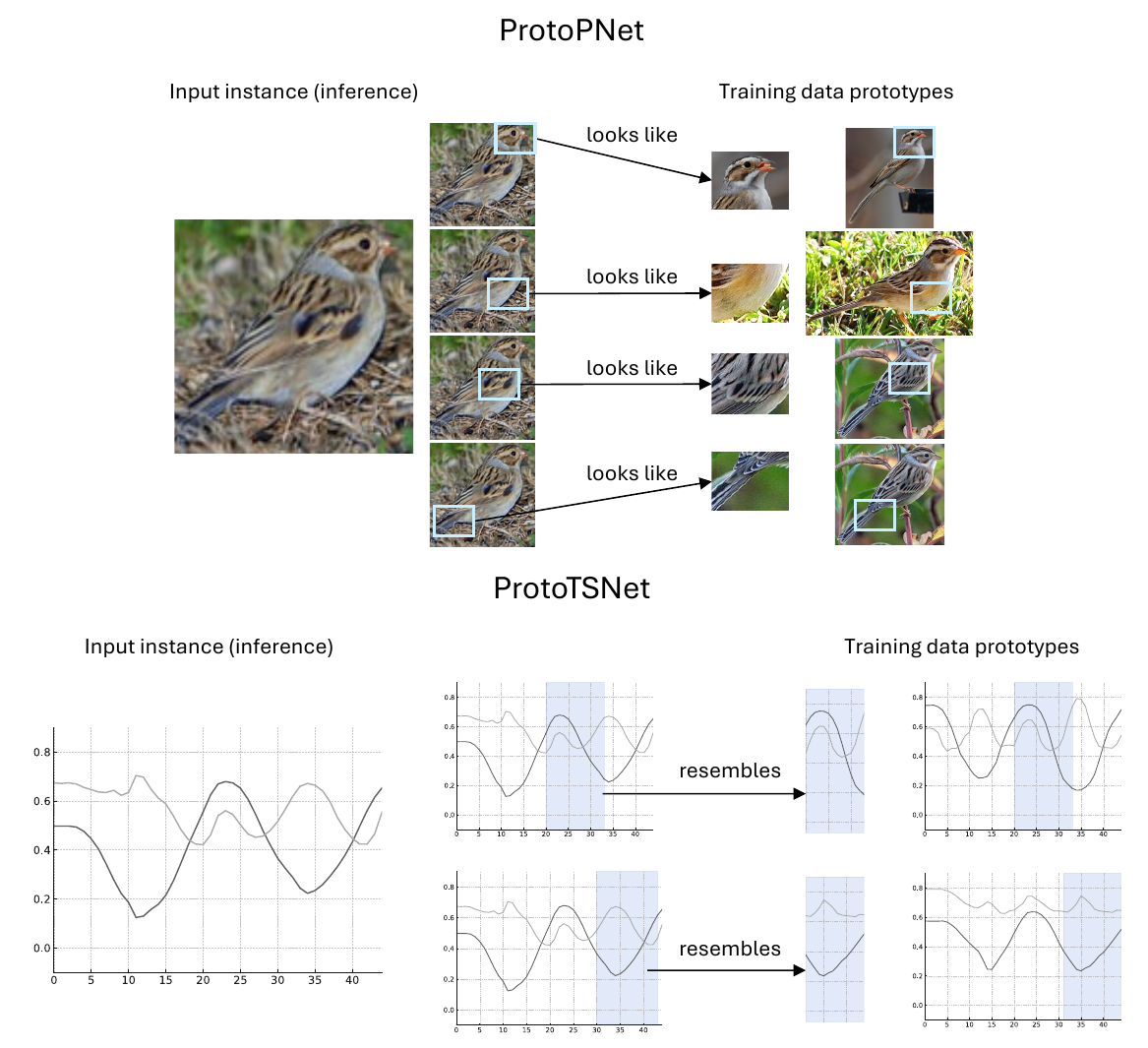}
    \caption{The general idea of how ProtoPNet and ProtoTSNet classify instances. In the case of ProtoPNet, it can be read as "this patch from the input image looks like that prototypical part from the training set." In the case of ProtoTSNet, it is "this subsequence from the input instance resembles that prototypical part from the training set." The ProtoPNet part of the image is sourced from its authors~\citep{chen2019looks}.}
    \label{fig:prototsnet_general_idea}
\end{figure}

Our work introduces a neural network architecture for Interpretable Time Series Classification with Prototypical Parts (ProtoTSNet) that leverages the principles of the ProtoPNet architecture and spans to univariate and multivariate time-series datasets. 
Although there have been similar endeavors to apply ProtoPNet's architecture or its foundational ideas~\citep{li2018deep} to time series analysis~\citep{ming2019prosenet,gee2019explaining}, they are limited to univariate series, their effectiveness is not tested on a wider range of problems, and/or source code is unavailable. 
We present a unique take on this problem, as our architecture differs from the mentioned works by introducing a custom encoder and training process modified to capture the prototypical parts in time-series.
The general idea behind ProtoTSNet is shown in Figure~\ref{fig:prototsnet_general_idea}.

Notably, our work addresses a relatively unexplored area in the literature by offering an ante hoc explainable approach for multivariate time series data.
Moreover, in the process of comparison with the selected methods in the literature, we rerun experiments to verify their ability to achieve the reported results. 
Based on the benchmarks we conducted on publicly available datasets, our method outperforms state-of-the-art explainable methods. Recognizing the importance of reproducibility, we ensure the accessibility of our method by publishing its source code\footnote{The code is published at \url{https://github.com/bmalkus/ProtoTSNet}}.

The remainder of the paper is organized as follows.
In Section~\ref{related-works} we present an overview of the state-of-the-art methods in the area of explainable AI for time series.
In Section~\ref{preliminaries} we state the problem and introduce the mathematical basis to describe our approach in detail in Section~\ref{prototsnet}.
We describe the evaluation and discuss the results in Section~\ref{eval}.
We conclude our paper in Section~\ref{summary}.

\section{Related Work}
\label{related-works}

In this section, we review recent advances in explainable time series classification, focusing on multivariate time series. The existing explainable methods can be categorized into ante-hoc methods, which incorporate interpretability directly into their architecture, and post-hoc methods, which apply explanatory techniques after model training.

\subsection{Post-hoc Explainable Methods}

Post-hoc methods include model-agnostic approaches such as SHAP~\citep{lundberg2017shap} and LIME~\citep{ribeiro2016lime}, and model-specific techniques like Grad-CAM~\citep{selvaraju2020gradcam}. In time series classification, Grad-CAM has been adapted to identify both relevant temporal segments and significant features, though this requires a specific architectural design.

MTEX-CNN~\citep{assaf2019mtex}, designed for multivariate time series, employs a two-stage architecture. The first stage processes input data of size $(T \times D)$, where $T$ denotes the temporal dimension and $D$ the number of features, using two-dimensional convolutions with $(k \times 1)$ kernels, independently processing each feature while retaining feature count and increasing channels. The second stage applies one-dimensional convolutions with $(k \times D)$ kernels temporally, followed by dense layers for classification. Grad-CAM is then applied to intermediate layers of both stages to generate explanations, highlighting feature contributions in the first stage and important temporal segments in the second. The MTEX-CNN method's reproducibility is limited, as no implementation was provided.

XCM~\citep{fauvel2021xcm}, another multivariate approach, extracts temporal and feature relevance in parallel rather than sequentially. It processes $(T \times D)$ input through two modules: one using $(k \times 1)$ kernels for feature processing and another applying one-dimensional convolutions with $(k \times D)$ kernels temporally. The outputs are combined and processed through additional convolutions and dense layers. The authors provided source code for both XCM and MTEX-CNN implementations, although our experiments did not replicate their reported performance.

LITEMVTime~\citep{ismail2025lite} is a post-hoc explainable method that applies Class Activation Maps (CAMs) to provide interpretability for multivariate time series classification. Built on the LITE architecture, which uses depthwise separable convolutions and boosting techniques, LITEMVTime generates temporal importance scores indicating which time stamps contribute the most to classification decisions. The~method has been demonstrated on human rehabilitation exercise evaluation, showing both competitive performance and explainability.

A more recent alternative to Grad-CAM, Dimension-wise Class Activation Mapping (dCAM)~\citep{boniol2022dcam}, specifically designed for multivariate time series, examines feature contributions directly through gradient analysis. Unlike Grad-CAM's focus on feature maps, dCAM generates temporal importance scores for each dimension, providing insights into feature interactions and temporal evolution.

\subsection{Ante-hoc Explainable Methods}

Ante-hoc methods include traditional interpretable models like decision trees or rule-based systems and modern neural network-based solutions. PETSC~\citep{feremans2022petsc} transforms time series using Symbolic Aggregate Approximation (SAX)~\citep{lin2003sax} to identify frequent sequential patterns of varying lengths. Although it supports both univariate and multivariate data, its explainability is limited to univariate cases. The method includes publicly available source code and performs well compared to other interpretable models.

LAXCAT~\citep{hsieh2021laxcat}, designed for multivariate time series, combines convolutional processing with dual attention mechanisms: variable attention for key features and temporal attention for critical intervals. While providing instance-level explanations, the lack of source code limits reproducibility.

MrSEQL~\citep{le2019mrseql}, limited to univariate data, combines SAX and Symbolic Fourier Approximation (SFA) representations, using an extended Sequence Learner to mine discriminative patterns. The authors provide the source code and results of the evaluation on the UCR Time Series Archive.

The r-STSF~\citep{cabello2024rstsf} method calculates aggregate statistics for subseries across multiple representations (frequency domain, derivative, and autoregressive), constructing a forest of randomized binary trees. While efficient and interpretable, it currently handles only univariate data. The method includes complete source code implementation.

\subsection{Prototype-based time series classification}

While prototype-based methods could be categorized under both ante-hoc and post-hoc approaches, we present them separately due to their particular relevance to our proposed method. These methods learn representative patterns from training data - either complete sequences or discriminative fragments - that serve as references for classification decisions. While they can provide interpretable insights through prototype comparison, the explainability aspect is not always explored in existing approaches. Although extensively studied in computer vision~\citep{salahuddin2022xai-med-review,poche2023example-based-survey}, applying prototype-based methods to time series data presents unique challenges, particularly in handling temporal dependencies and varying sequence lengths.

ProtoFac~\citep{das2020protofac} is a post-hoc method supporting multivariate time series. It replaces selected neural network layers with a prototype matrix through explainable factorization, constraining prototypes to realistic data segments while maintaining model fidelity. The method decomposes latent activations into interpretable prototypes and corresponding weights after model training, though lack of source code limits reproduction.

ProSeNet~\citep{ming2019prosenet} takes an ante-hoc approach for univariate data, incorporating interpretability directly by learning whole-instance prototypes from RNN latent space during training. The method optimizes for simplicity, diversity, and sparsity, and allows domain experts to refine prototypes without altering the model structure, but no implementation was provided.

Gee et al.~\citep{gee2019explaining} present another ante-hoc approach focused exclusively on univariate time series. Their autoencoder-prototype architecture learns prototypes directly from the latent space during training, with a diversity penalty ensuring representative coverage of the data distribution. The method lacks publicly available implementation.

DPSN~\citep{tang2020dpsn} is an ante-hoc method that combines "representative shapelets" (whole instances as prototypes) with "discriminative shapelets" (differentiating sub-sequences) for few-shot learning. The method uses SFA features and metric learning. The source code is provided, but the method remains limited to univariate data.

\begin{table}[ht]
    \caption{Comparison of state of the art methods from the literature with ours.}
    \label{tab:sota_comparison}
    \setlength{\tabcolsep}{5pt}
    \centering
        \begin{tabular}{l|c|c|c}
        \hline
        Method                            &  Multivariate  &  Ante-hoc explainable  &  Source code provided  \\
        \hline
        \hline
        MTEX-CNN \citep{assaf2019mtex}     &  \checkmark    &                           &  \footnotemark[1]                    \\
        XCM \citep{fauvel2021xcm}          &  \checkmark    &                           &  \checkmark            \\
        \makecell{LITEMVTime \\ \citep{ismail2025lite}}  & \checkmark     &                           &  \checkmark            \\
        PETSC \citep{feremans2022petsc}    &  \checkmark    &  \footnotemark[2]                       &  \checkmark            \\
        LAXCAT \citep{hsieh2021laxcat}     &  \checkmark    &  \checkmark               &                        \\
        MrSEQL \citep{le2019mrseql}        &                &  \checkmark               &  \checkmark            \\
        r-STSF \citep{cabello2024rstsf}    &                &  \checkmark               &  \checkmark            \\
        ProtoFac \citep{das2020protofac}   &  \checkmark    &                           &                        \\
        ProSeNet \citep{ming2019prosenet}  &                &  \checkmark               &                        \\
        - \citep{gee2019explaining}        &                &  \checkmark               &                        \\
        DPSN \citep{tang2020dpsn}          &                &  \checkmark               &  \checkmark            \\
        \textbf{ProtoTSNet}                &  \checkmark    &  \checkmark               &  \checkmark            \\
        \hline
        \end{tabular}
    \footnotetext[1]{Implementation was not provided by the original authors, but XCM authors implemented this method}
    \footnotetext[2]{Explainability limited to univariate data}
\end{table}

\section{Preliminaries}
\label{preliminaries}
In this section, we outline the core objectives and challenges that our research addresses in the context of explainable time series classification. We define the problem of multivariate time series classification, discuss feature importance in time series analysis, and then discuss the prototype term.

\subsection{Multivariate Time Series Classification}

A multivariate time series $X$ is represented as a matrix $X \in \mathbb{R}^{d \times T}$, where $d$ denotes the number of dimensions (features) and $T$ denotes the length of the time series. Each feature (dimension) of $X$ can be represented as $X_i$, where $i = 1, 2, \ldots, d$, and each $X_i$ is a sequence of observations over time $t = 1, 2, \ldots, T$.

The classification task involves the construction of a function $f: \mathbb{R}^{d \times T} \mapsto \{1, 2, \ldots, C\}$ that maps a multivariate time series $X$ to a class label\linebreak $y \in \{1, 2, \ldots, C\}$, where $C$ is the number of classes. This function $f$ is learned from a training set $\left\{(X^{(1)}, y^{(1)}), (X^{(2)}, y^{(2)}), \ldots, (X^{(n)}, y^{(n)})\right\}$ consisting of $n$ pairs of multivariate time series and their corresponding class labels. The challenge in MTSC is to effectively capture both the temporal dynamics within each feature $X_i$ and the interactions between different features to make accurate predictions about the class label~$y$.

\subsection{Feature Importance in Time Series Analysis}

Understanding the importance of features in time series classification is essential both for model interpretation and effectiveness. In this context, feature importance refers to the relative significance of each feature (or dimension) in contributing to the model's predictive performance. Identifying key features helps in uncovering the underlying patterns and relationships that drive the prediction outcomes.

Mathematically, let us denote the feature importance vector as $I \in \mathbb{R}^d$, where each element $I_i$ represents the importance of the $i$-th feature in the classification model. The aim is to evaluate $I$ in a way that highlights the features that have a significant impact on the model's decisions.

\subsection{Prototypes in Time Series Analysis}

The term "prototype" in time series classification has varied interpretations across the literature. Some studies consider prototypes as representative full-length time series~\citep{zhang2020tapnet,gee2019explaining}, while others treat them as subsequences, either proper, maintaining temporal order \citep{das2020protofac}, or improper, capturing similar patterns across different series~\citep{tang2020dpsn,ming2019prosenet}.

In our work, we define prototypical parts as proper subsequences of multivariate time series. Formally, for an input series $X \in \mathbb{R}^{d \times T}$, a prototypical part $P = X_{*,[t_s:t_e]}$ spans all features ($*$) over a temporal segment $[t_s:t_e]$, where $1 \leqslant t_s < t_e \leqslant T$. We denote the length of this subsequence by $L$, where $L = t_e - t_s + 1$. These prototypes exist in both the input space ($P$) and the encoded space ($P_l \in \mathbb{R}^{l \times L}$), with mappings between them. Using proper subsequences, our approach introduces higher granularity and flexibility into the classification process, improving both interpretability and performance by focusing on the most informative segments of the data through prototype identification and feature importance calculation.

\section{Prototypical Parts for Multivariate Time-series}
\label{prototsnet}
In this section, we introduce our proposed architecture designed for multivariate time series. We discuss the general principle of operation, the proposed encoder architecture, and the calculation of the feature importance.

\begin{figure}
    \centering
    \includegraphics[width=\textwidth]{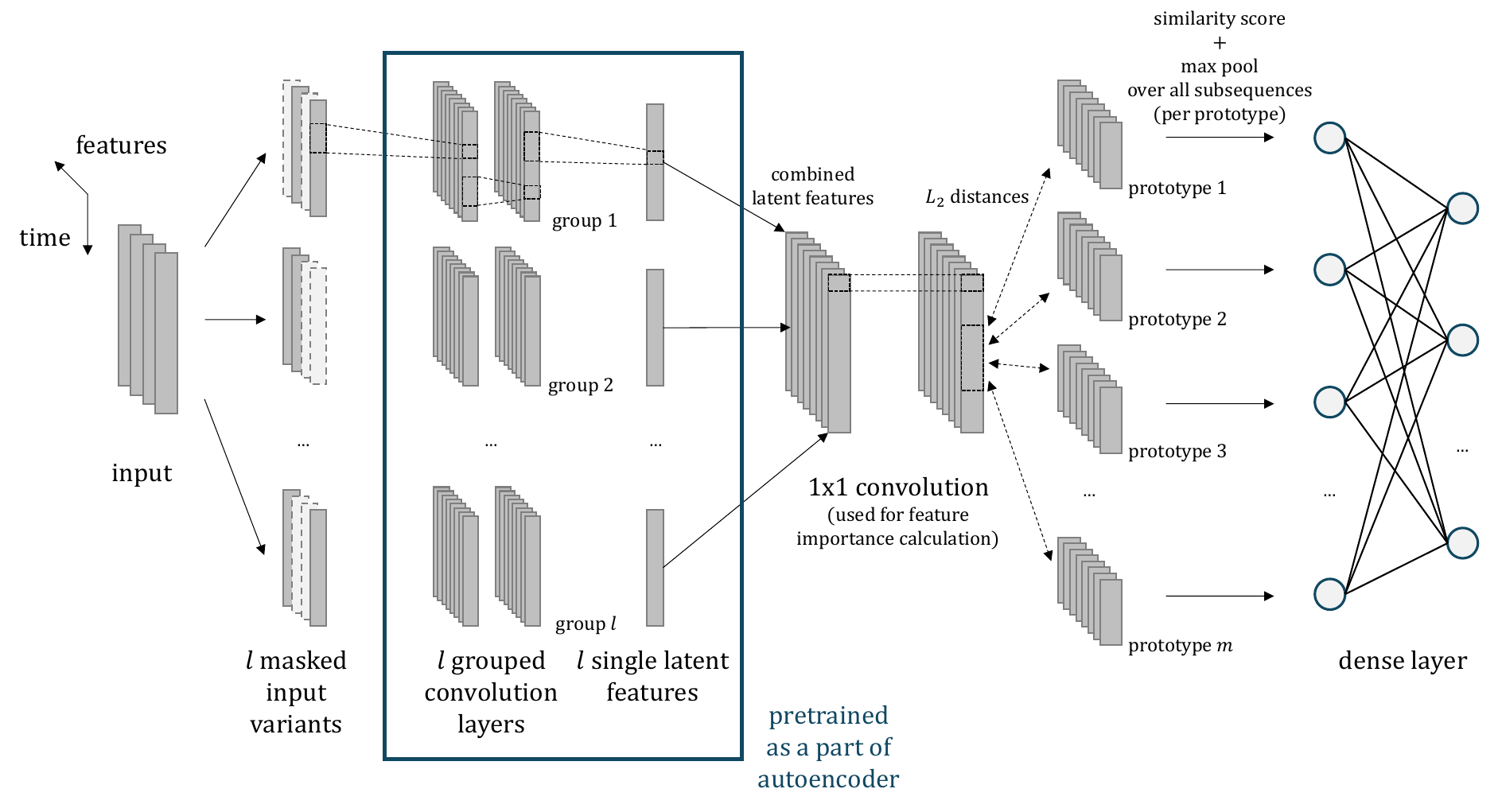}
    \caption{Model Architecture. The input multivariate time series is processed through $l$ binary masks, each preserving a different subset of features controlled by reception parameter $r$. These masked variants are fed into a grouped convolution encoder ($l$ groups), where each group processes a distinct feature subset independently, producing a single latent feature per group. The encoder outputs are mixed through a $1 \times 1$ convolution layer, whose weights enable feature importance calculation. The prototype layer then computes $L_2$ distances between latent subsequences and learned prototypes, applying max pooling over all subsequences to obtain similarity scores per prototype. Finally, these similarity scores are processed by a dense layer to produce class predictions.}
    \label{fig:prototsnet_arch}
\end{figure}

\subsection{Model Architecture}

Our approach draws inspiration from the architecture of ProtoPNet~\citep{chen2019looks}, but introduces several key modifications, specifically designed to handle multivariate time series data. At its core, the architecture implements a feature masking mechanism combined with grouped convolutions, enabling both effective feature representation learning and interpretability through feature importance calculations. The~complete architecture is illustrated in Figure~\ref{fig:prototsnet_arch}.

\subsubsection{Input Processing and Feature Masking}

The first distinctive element of our architecture is the introduction of a feature masking mechanism. Given an input multivariate time series with $d$ features, we generate $l$ binary masks, where $l$ represents the number of encoder groups (typically $l = 32$ in our experiments). Each mask has a length equal to the number of input features ($d$).

For the $i$-th encoder group, each mask element $\delta_{im} \in \{0, 1\}$ determines whether the $m$-th input feature is retained (1) or masked out (0). The proportion of retained features is controlled by a parameter $r \in (0, 1)$, called the \textit{reception parameter}, which determines how many features each group receives. Specifically, each mask retains exactly $\lfloor r \cdot d \rfloor$ randomly selected features.

For example, with $d = 10$ input features, $l = 4$ groups, and $r = 0.5$, we generate 4 binary masks, each retaining 5 randomly selected features from the original 10. This process yields $l$ variants of the input time series, each preserving a different subset of features, where each subset contains exactly $\lfloor r \cdot d \rfloor$ features.

The parameters $l$ and $r$ serve complementary roles: $l$ determines the number of parallel processing groups (and consequently the dimensionality of the latent prototype space), while $r$ controls the sparsity of features within each group, affecting both the~computational efficiency and the quality of the calculation of features importance.

The goal of this masking process is to prevent the mixing of information between dimensions that carry significant information (prototypical parts) and those that do not (e.g., dimensions considered as noise). 
In the case of image modalities, all channels (e.g., RGBA) are typically considered important for building a prototype. 
Conversely, in time series data, the channels (or dimensions) might be less tightly connected. 
As a result, separating these dimensions during the prototype selection process plays a~crucial role in ensuring the effectiveness of the approach.

\subsubsection{Grouped Convolution Encoder}
\label{sec:group_conv_enc}

The masked input variants are processed through a convolutional encoder that employs group convolution with $l$ groups. Given an input time series $X \in \mathbb{R}^{d \times T}$, after feature masking, we obtain $l$ masked variants, each of shape $\mathbb{R}^{d \times T}$ (with $(1-r) \cdot d$ features zeroed out).

The grouped convolution encoder processes these variants with the following shape transformations:
\begin{itemize}
    \item \textbf{Input:} $l$ masked variants of shape $[batch\_size, d, T]$
    \item \textbf{Grouped Convolution Layers:} Each group processes its corresponding masked input independently. Each group produces a single latent feature of shape $[batch\_size, 1, T]$.
    \item \textbf{Single Latent Features:} Since each group outputs exactly one feature, we refer to these as "single latent features." The combined outputs (latent features) from all $l$ groups have the shape $[batch\_size, l, T]$.
\end{itemize}

Unlike standard convolution layers where all inputs are connected to all outputs, group convolution restricts connections to occur only within designated groups. By~setting the number of groups to $l$ in every group convolution layer, each masked input variant is processed by an independent group, effectively serving as a mini-encoder, preventing information mixing between different feature subsets.

\subsubsection{Feature Mixing and Prototype Space}
\label{sec:feat_mixing_proto_space}

The separated feature representations from the grouped encoder (shape\linebreak $[batch\_size, l, T]$) are subsequently processed through a $1 \times 1$ convolution layer. This~layer maintains the shape.

The primary purpose of this layer is to perform a controlled mixing of information from different feature subsets, enabling the calculation of feature importance through analysis of the mixing weights (detailed in Section~\ref{sec:feat_imp_calc} below). The output of this layer forms the \textit{prototype latent space} where the prototype distance calculations are performed.

The final shape entering the prototype layer is $[batch\_size, l, T]$, where subsequences of length $L$ are extracted for prototype matching.

\subsubsection{Prototype Layer and Classification}

The prototype layer computes similarities between learned prototypes and subsequences in the latent space. Given the latent representation $Z \in \mathbb{R}^{l \times T}$ from the $1 \times 1$ convolution layer (Section~\ref{sec:feat_mixing_proto_space}), we extract all possible subsequences of length $L$ along the temporal dimension.

Specifically, for each prototype $P_l^{(j)} \in \mathbb{R}^{l \times L}$ from the set of $m$ prototypes and all subsequences $\tilde{Z} \in \mathbb{R}^{l \times L}$ of length $L$ in the latent series $Z$, the layer calculates $L_2$ distances and transforms them into similarity scores. This transformation is formalized~as:

\begin{equation}
    \label{eq:similarity}
    sim(P_l^{(j)}, Z) = \max_{\tilde{Z} \in subseq(Z)} \log{\left(\frac{{\left\Vert \tilde{Z} - P_l^{(j)} \right\Vert}_2^2 + 1}{{\left\Vert \tilde{Z} - P_l^{(j)} \right\Vert}_2^2 + \epsilon}\right)}
\end{equation}

where $\epsilon$ is a small constant that prevents division by zero, and $\text{subseq}(Z)$ denotes the set of all subsequences of length $L$ extracted from $Z$.

The layer focuses on the most significant match within the series by selecting the~maximum similarity score for each prototype across all possible subsequences. This max-pooling operation ensures that each prototype activation represents the strongest match found anywhere in the input time series.

These maximum similarity scores serve as prototype activations, which are then transformed into class predictions through a dense layer performing linear combination. The dense layer learns which prototypes are the most indicative of each class, enabling the final classification decision.

Since the temporal structure is preserved in the latent space, prototypes can be directly mapped back to segments in the input space, taking into account the receptive field. This direct mapping enables a straightforward interpretation of the learned representations, allowing domain experts to understand which temporal patterns the~model considers important for classification.

\subsection{Feature Importance Calculation}
\label{sec:feat_imp_calc}

The proposed architecture enables direct calculation of feature importance by leveraging its controlled feature mixing mechanism. The key aspect of this calculation lies in the weights of the $1 \times 1$ convolution layer that follows the grouped convolution encoder structure. To calculate the importance of a given input feature, we analyze how it influences each feature in the prototype space through the network's connectivity pattern. For the $m$-th input feature, its contribution is determined by two factors:
\begin{itemize}
    \item the binary masks that control which encoders receive this feature,
    \item the weights in the $1 \times 1$ convolution layer that mix the encoder outputs.
\end{itemize}
Formally, the importance $I_m$ of the $m$-th input feature is calculated as:
\begin{equation}
    I_m = \sum_{j=1}^{l} \left( \left\Vert \sum_{i=1}^{l} \delta_{im} w_{ij} \right\Vert \right)
\end{equation}
where:
\begin{itemize}
    \item $i$ indexes the encoder outputs (inputs to the convolution layer),
    \item $j$ indexes the features in the prototype space (outputs of the convolution layer)
    \item $\delta_{im} \in {0,1}$ indicates whether the $m$-th feature is included in the $i$-th encoder's input
    \item $w_{ij}$ represents the weight connecting the $i$-th encoder output to the $j$-th prototype space feature
\end{itemize}

This calculation first determines the contribution of feature $m$ to each prototype space feature $j$ by summing the relevant pathway weights ($\delta_{im} w_{ij}$). The absolute values of these contributions are then summed across all prototype space features to obtain the overall importance score. This approach ensures that both positive and negative influences are captured in the final importance measure.

There are two important considerations that affect the feature importance calculations. First, the method produces global feature importance scores that characterize the overall influence of each feature across all prototypes, rather than prototype-specific importance values. Second, the reliability of these importance scores is tied to the encoder's reception parameter~$r$. This parameter creates a trade-off: larger $r$ values allow more features to be processed simultaneously by each encoder, potentially leading to importance score overestimation due to increased feature interaction, while smaller $r$ values maintain feature separation but may miss some of the contextual relationships.

\begin{figure}[t]
    \centering
    \subfloat{
        \centering
        \begin{minipage}[c][\height][c]{0.48\textwidth}
            \includegraphics[width=\textwidth]{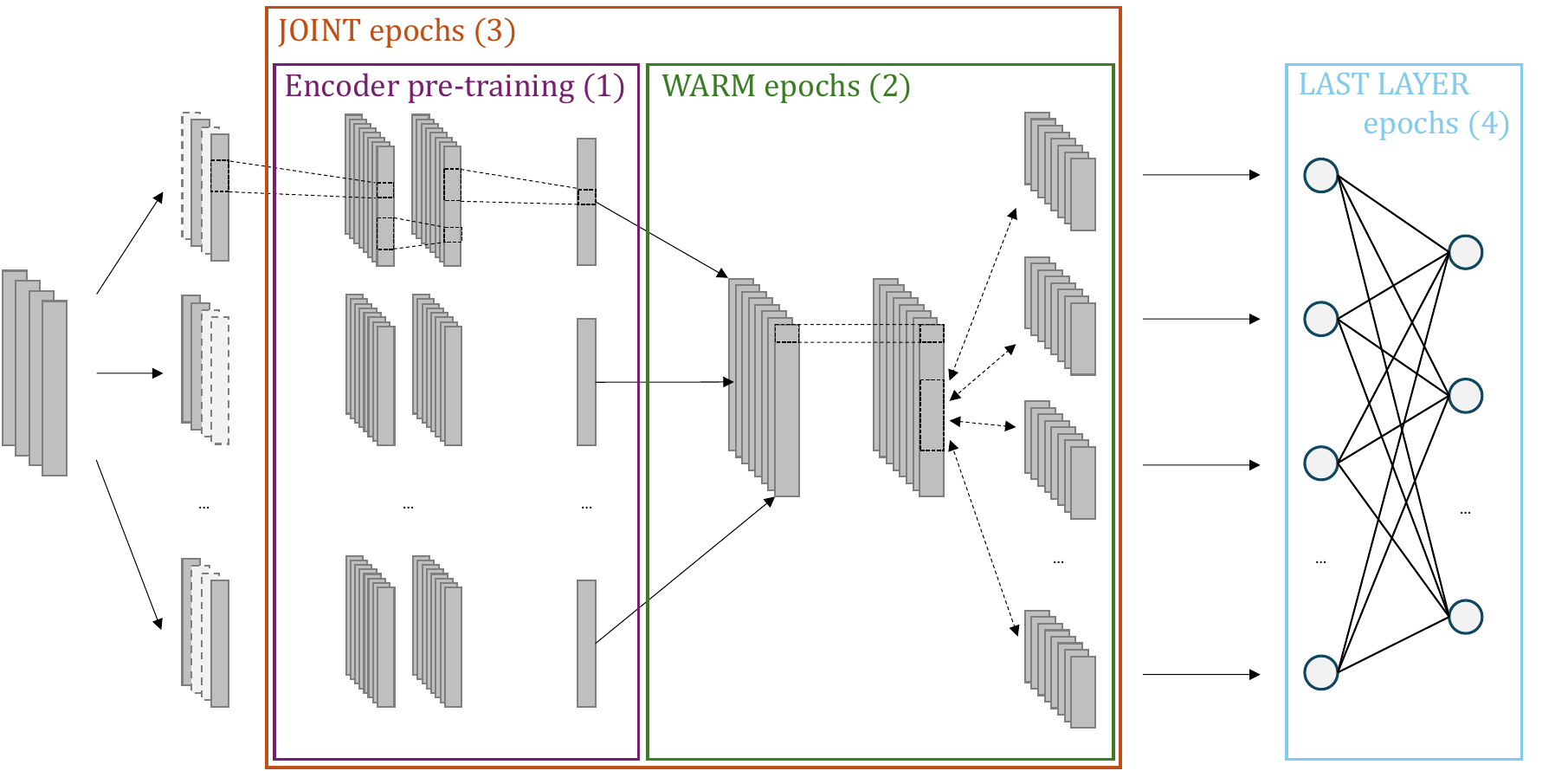}
        \end{minipage}
    }
    \subfloat{
        \centering
        \begin{minipage}[c][\height][c]{0.48\textwidth}
            \includegraphics[width=\textwidth]{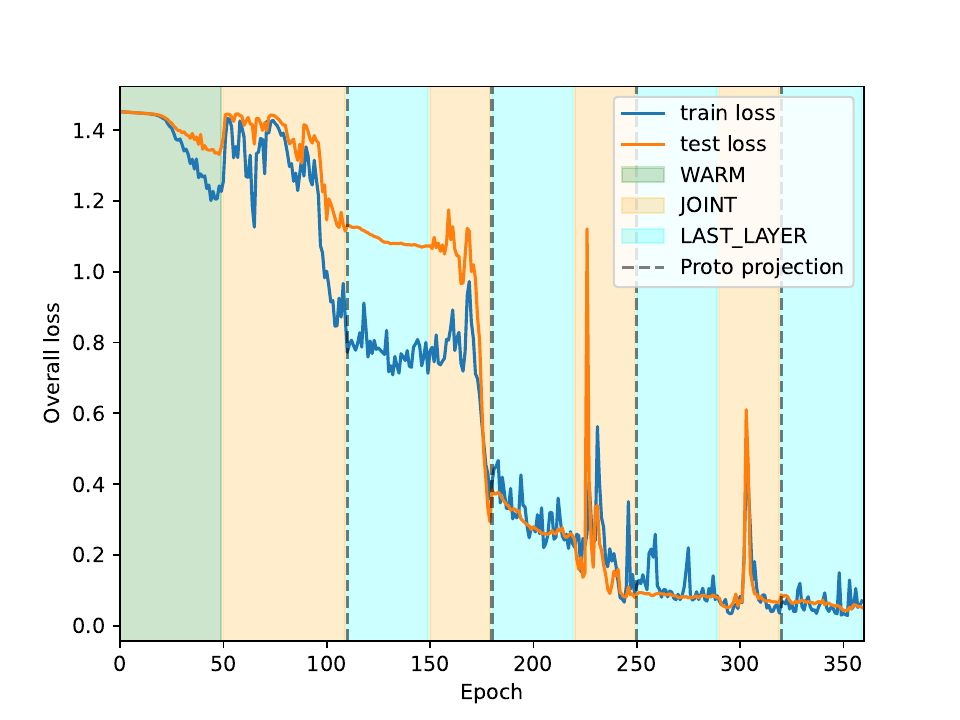}
        \end{minipage}
    }
    \caption{The training process of ProtoTSNet and its different stages mapped to the learning curve. Numbers in brackets represent order of the training stage.}
    \label{fig:training_process}
\end{figure}

\subsection{Training Process}

The complete training process consists of two distinct phases: encoder pretraining and main ProtoTSNet training, as illustrated in Figure~\ref{fig:training_process} and detailed in Algorithm~\ref{algo:training_process}.

\textbf{Phase 1: Encoder Pretraining.} We train a standard autoencoder to reconstruct the input time series using MSE loss. No prototypes are involved - the goal is to learn meaningful feature representations. After pretraining, the decoder is discarded, and only the encoder weights are retained for initialization of the main model.

\textbf{Phase 2: Main ProtoTSNet Training.} We introduce the prototypes as learnable parameters and train the complete architecture. The pretrained encoder weights serve as initialization, while prototypes are randomly initialized in the latent space.
The final classification layer is initialized to promote early prototype-class alignment:
weights corresponding to a prototype’s own class (positive connections) are set to~1,
weights to other classes (negative connections) to~-0.5, and all biases are disabled
in both the final layer and the $1\times1$ feature-mixing convolution layer.
This initialization provides a meaningful separation between supportive and opposing
prototype evidence from the start, reinforcing the interpretability of the model.
The training follows a cyclical process consisting of four stages, repeated multiple times:

\begin{enumerate}
    \item \textbf{Warm epochs:} SGD optimization applied exclusively to prototypes and the $1 \times 1$ convolution layer weights, while encoder parameters remain frozen. This allows prototypes to adapt to the pretrained latent space.
    
    \item \textbf{Joint epochs:} SGD optimization extended to include all parameters (prototypes, $1 \times 1$ convolution layer, and encoder). An exponential decay cyclic learning rate scheduler is employed to enhance exploration of the prototype space.
    
    \item \textbf{Prototype projection:} Each prototype is projected onto the closest latent patch from the training data, ensuring that prototypes correspond to actual patterns observed in the training set. This step is crucial for maintaining the "this resembles that" interpretability property.
    
    \item \textbf{Last layer optimization:} Only the final dense layer is optimized while all other parameters remain frozen, allowing the model to learn optimal prototype-to-class mappings.
\end{enumerate}

The model's training is guided by different loss functions depending on the training stage:

\textbf{Warm and Joint Epochs Loss:}
\begin{equation}
\mathcal{L}_{main} = \mathcal{L}_{ce} + \lambda_{clst} \mathcal{L}_{clst} + \lambda_{sep} \mathcal{L}_{sep} + \lambda_{conv} \mathcal{L}_{l1\_conv}
\end{equation}

\textbf{Last Layer Optimization Loss:}
\begin{equation}
\mathcal{L}_{last} = \mathcal{L}_{ce} + \lambda_{last} \mathcal{L}_{l1\_last}
\end{equation}

where:
\begin{enumerate}
    \item $\mathcal{L}_{ce}$ is the cross-entropy loss for classification accuracy.
    \item $\mathcal{L}_{clst}$ (clustering regularization) is defined as:
    \begin{equation}
    \mathcal{L}_{clst} = \frac{1}{n} \sum_{i=1}^{n} \min_{j:P_l^{(j)} \in \mathcal{P}_{y_i}} \min_{\tilde{Z} \in \text{subseq}(Z_i)} \|\tilde{Z} - P_l^{(j)}\|_2^2
    \end{equation}
    which encourages each training series to have at least one latent subsequence close to a prototype of its class.
    \item $\mathcal{L}_{sep}$ (separation regularization) is defined as:
    \begin{equation}
    \mathcal{L}_{sep} = -\frac{1}{n} \sum_{i=1}^{n} \min_{j:P_l^{(j)} \notin \mathcal{P}_{y_i}} \min_{\tilde{Z} \in \text{subseq}(Z_i)} \|\tilde{Z} - P_l^{(j)}\|_2^2
    \end{equation}
    which encourages latent subsequences to remain distant from prototypes of other classes.
    \item $\mathcal{L}_{l1\_conv}$ applies L1 sparsity regularization to the $1 \times 1$ convolution layer weights, helping identify important features.
    \item $\mathcal{L}_{l1\_last}$ applies L1 regularization to negative weights in the last layer. Since the last layer connects prototype activations to class predictions, negative weights represent "negative evidence" (e.g., "this is class 1 because it does NOT resemble prototype of class 2"). By penalizing negative weights, we encourage the model to rely on positive prototype evidence for classification, improving interpretability.
\end{enumerate}

Here, $\mathcal{P}_{y_i}$ denotes the set of prototypes belonging to class $y_i$, $Z_i$ is the latent representation of the $i$-th training series, and $\text{subseq}(Z_i)$ represents all subsequences of length $L$ extracted from $Z_i$. The clustering and separation terms shape the latent space into a semantically meaningful structure, while the last layer optimization ensures sparsity in cross-class prototype connections, avoiding negative reasoning patterns.

\begin{algorithm}[t]
\caption{Simplified flow of the training process}
\label{algo:training_process}
\begin{algorithmic}[1]
    \Require
        \Statex $E_P :=$ \textit{number of pretraining epochs}
        \Statex $E_W :=$ \textit{number of warm epochs}
        \Statex $E_L :=$ \textit{last layer optimization epochs}
        \Statex $I_P :=$ \textit{number of joint epochs between pushes}
    \Statex

    \For {$epoch := 1$ to $E_P$}
        \State \textit{Pretrain encoder on training data (as part of autoencoder)}
    \EndFor
    \For {$epoch := 1$ to $E_W$}
        \State // Warm epoch
        \State \textit{Train prototypes layer only}
    \EndFor
    \While {\textit{end condition not met}}
        \For {$epoch := 1$ to $I_P$}
            \State // Joint epoch
            \State \textit{Train prototypes layer along with encoder}
        \EndFor
        \State // Push
        \State \textit{Project each prototype onto closest latent patch in training data}
        \For {$epoch := 1$ to $E_L$}
            \State // Last layer epoch
            \State \textit{Train only the last layer}
        \EndFor
    \EndWhile
\end{algorithmic}
\end{algorithm}

\section{Experimental evaluation}
\label{eval}
In this section, we describe experimental evaluation of ProtoTSNet.
We start with an~experiment conducted on a synthetic dataset designed to test our model's capability in identifying regions of importance within time series data and to calculate feature importance.
We then move on to the suite of experiments conducted on the UEA datasets archive\footnote{See \url{https://timeseriesclassification.com}.}. This evaluation allows us to compare our model against state of the~art models in the field of time series classification.

\subsection{Synthetic Dataset}

Our synthetic dataset consists of three features, of which two are significant for class distinction and one is not. It contains four classes, each marked by distinct saw and rectangular patterns in significant features. A small amount of white noise is introduced into the series. Significant parts are limited to the initial 40 time steps, the~subsequent 60 are randomized and independent from the classes. The dataset was generated with 1000 training examples and 100 test examples.

\begin{figure}
    \centering
    \includegraphics[width=\textwidth]{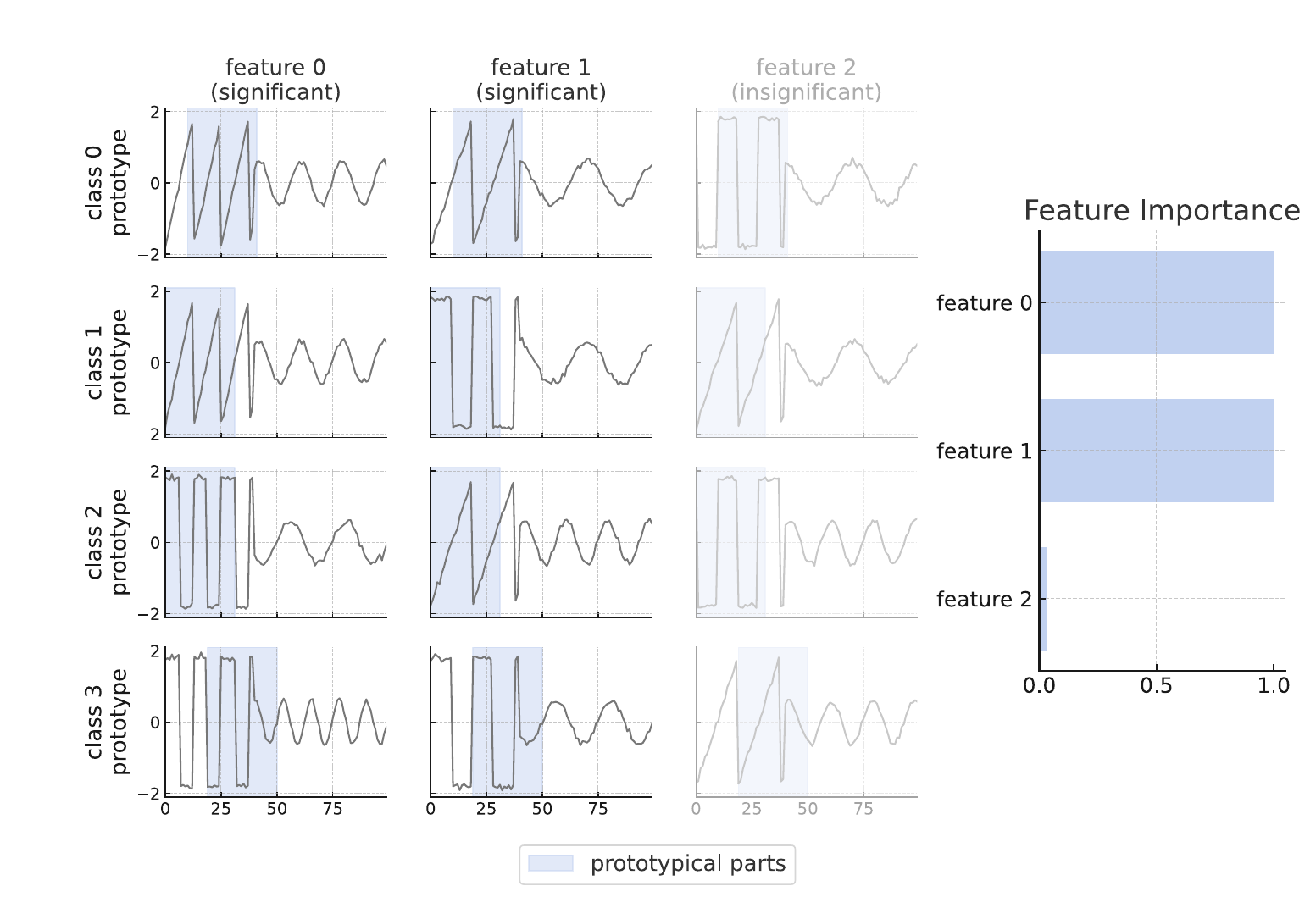}
    \caption{Prototypical parts (blue background) learned on our synthetic dataset and calculated feature importance. 
    Features are split into separate plots for clarity (each row represents a single prototypical part). 
    Each prototype correctly captures the significant part (initial 40 time steps), however, some of them cover the insignificant area as well. 
    This is specific to the design of our model and the experiment setup since capturing a small portion of the significant part is enough to distinguish the classes. 
    There are three features, of which two are significant for the class distinction and one is not. 
    The feature importance calculated reflects this, and the insignificant feature is dimmed out.
    }
    \label{fig:artificial_protos}
\end{figure}

The model was trained for four cycles of (joint epochs → prototype projection → last layer optimization), with reception of 0.75, prototype length of 20, and one prototype per class. The model achieved 100\% classification accuracy as expected for this simple dataset. Figure \ref{fig:artificial_protos} presents the learned prototypes and calculated feature importance score.

The prototypes have correctly captured the significant fragments of the input across all classes, demonstrating the model's capability to identify key temporal patterns. The coverage of significant parts varies between classes - for example, in class 3, the~prototype includes a substantial portion of insignificant area. This stems from our model's design, where prototype activation is driven by similarity measures, allowing partial matches with class-specific segments to cause significant activation. In real-world scenarios, where prototypes are typically imperfect and insignificant parts vary between inputs, the model tends to focus more on capturing significant segments.

\subsection{UEA Datasets}
\label{sec:uea_datasets}

We evaluated our model using the UEA multivariate time series repository~\citep{bagnall2018uea}, which contains 30 data sets that span various domains, dimensions, and complexity levels. These datasets show significant variety in their characteristics: the~number of classes ranges from 2 to 39, the number of features from 2 to 1345, and the sequence length from 8 to 17984. Each dataset comes with predefined train/test splits. The complete dataset characteristics and corresponding optimal hyperparameters are presented in Table \ref{tab:datasets_characteristics} in the Appendix.

We evaluated ProtoTSNet against both explainable and non-explainable methods:
\begin{itemize}
    \item \textbf{Ante-hoc explainable:}
    PETSC~\citep{feremans2022petsc}, and a Shapelet-based method implemented using the \texttt{ShapeletTransformClassifier}
    from the \texttt{sktime} library (v0.27.0)~\citep{loning2019sktime}. This implementation follows the
    binary shapelet transform approach of~\citep{hills2014shaplet}, using random shapelet sampling
    and the\linebreak \texttt{RotationForestClassifier} as the downstream estimator.
    Shapelets were extracted independently per channel (univariate shapelets per dimension).
    \item \textbf{Post-hoc explainable:} XCM~\citep{fauvel2021xcm}, MTEX-CNN~\citep{assaf2019mtex}, LITEMVTime~\citep{ismail2025lite}
    \item \textbf{Non-explainable:} ROCKET~\citep{dempster2020rocket} (state-of-the-art time series classification method),
    TapNet~\citep{zhang2020tapnet} (established non-explainable prototype-based method)
\end{itemize}

We excluded LAXCAT (Table~\ref{tab:sota_comparison}) from our comparison due to the lack of clarity regarding the exact preprocessing steps used in the original study, which were crucial to the overall training process. The authors did not include these details in their paper and did not provide a source code implementation.

We conducted all experiments using the original implementations and hyperparameters provided by their respective authors, with exceptions noted for MTEX-CNN (using XCM authors' implementation) and PETSC (modified to use 5-fold cross-validation instead of test set optimization).

For our method, we performed hyperparameter optimization through grid search with 5-fold cross-validation, exploring:
\begin{itemize}
    \item Reception parameter $r$: {0.25, 0.5, 0.75, 0.9}
    \item Prototypical parts length $L$: {1\%, 10\%, 25\%, 50\%, 100\%} of the time series length
\end{itemize}
All other hyperparameters remained constant across experiments, they are provided in Section~\ref{secA:common-params} in the~Appendix. Complete cross-validation results are available in our GitHub repository. The model training consisted of two phases:
\begin{itemize}
    \item Encoder pretraining: 50 epochs as part of an autoencoder with MSE loss
    \item Main training: four cycles of (joint epochs $\rightarrow$ prototype projection $\rightarrow$ last layer optimization), totaling approximately 360 epochs
\end{itemize}

The classification accuracy scores for all methods are presented in Table \ref{tab:accu_scores} in the~Appendix, with bold values indicating the best result for each dataset. Incomplete cases are marked as "N/A" in the table and represent trials that failed to complete due to resource constraints being exceeded. Summary statistics, including average ranks and the number of wins/ties, are provided at the bottom of the table. For the purpose of calculating average ranks, we assigned the lowest ranks to the "N/A" cases, as these represent scenarios requiring impractical computational resources.

ProtoTSNet achieves competitive performance across all method categories, ranking best among ante-hoc explainable methods with an average rank of 3.90. Among all methods, ROCKET achieves the best average rank of 2.73, followed by LITEMVTime (3.03). This performance hierarchy reflects the natural accuracy advantages of non-explainable methods and post-hoc explainable methods, which are not constrained by ante-hoc explainability requirements.

Performance analysis reveals dataset-specific patterns. ProtoTSNet demonstrates strong performance on datasets like DuckDuckGeese and HandMovementDirection, where neural network-based methods and ROCKET also perform well, suggesting these datasets benefit from convolutional filtering and temporal pattern extraction capabilities. On datasets like EthanolConcentration and Handwriting, shapelet-based and/or post-hoc methods show superior performance, which may indicate scenarios where global statistical features or fine-grained local patterns are more informative than the intermediate-length prototypical parts learned by our method.

\begin{figure}
    \centering
    \includegraphics[width=0.75\textwidth]{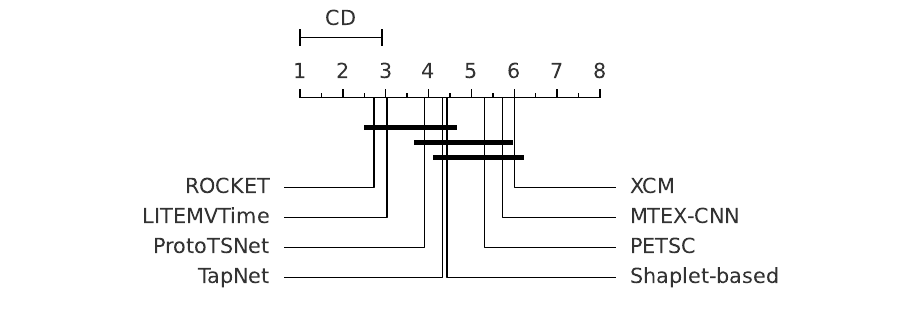}
    \caption{Critical difference diagram for methods evaluated on the UEA multivariate datasets ($\alpha = 0.05$). ProtoTSNet achieves competitive performance, ranking best among ante-hoc explainable methods (ProtoTSNet, PETSC, Shapelet-based) and showing comparable performance to non-explainable methods (ROCKET, TapNet) and post-hoc explainable method (LITEMVTime). Note that non-explainable methods and post-hoc explainable methods naturally have accuracy advantages as they are not constrained by ante-hoc explainability requirements.}
    \label{fig:cd_diag}
\end{figure}

We conducted a statistical analysis of our results using the critical difference test with Nemenyi post-hoc analysis~\citep{demsar2006statistical}. The resulting critical difference diagram, generated using the Orange data mining library~\citep{demsar2013orange} and presented in Figure \ref{fig:cd_diag}, compares the average ranks across all methods and datasets. In the diagram, methods connected by a horizontal bar do not exhibit statistically significant differences in performance. ProtoTSNet achieves competitive performance among all methods, ranking best among ante-hoc explainable methods and showing comparable performance to non-explainable methods and post-hoc explainable methods. Note that non-explainable methods and post-hoc explainable methods naturally have accuracy advantages as they are not constrained by ante-hoc explainability requirements.

\subsubsection{Prototypical Parts}

We use the Libras dataset~\citep{dias2009hand} to illustrate the interpretability of prototypical parts. Libras contains 15 classes with 24 instances per class (hand-movement trajectories).

\textbf{Setup for qualitative analysis.}
For the Libras dataset we used the original train/test split provided with the UEA archive.
Unless specified differently, all hyperparameters follow UEA dataset experiments, see Section~\ref{secA:common-params} in the Appendix.
For this qualitative analysis we fixed the number of prototypes to 2 per class (down from 10 in UEA dataset), and varied the last-layer L1 coefficient
$\lambda_{\text{last}} \in \{ 1\!\times\!10^{-3},\, 3\!\times\!10^{-3},\, 1\!\times\!10^{-2} \}$.
Using 2 prototypes per class slightly reduces accuracy but enables clearer visualization of class-specific patterns (quantified later in this subsection).
Numerical results for diagnostics and accuracy were aggregated over 5 independent runs for each $\lambda_{\text{last}}$.
Figures in the main text show the last run of the middle setting $\lambda_{\text{last}}=3\!\times\!10^{-3}$ (chosen arbitrarily without preference among runs).

\begin{figure}
    \centering
    \includegraphics[width=\textwidth]{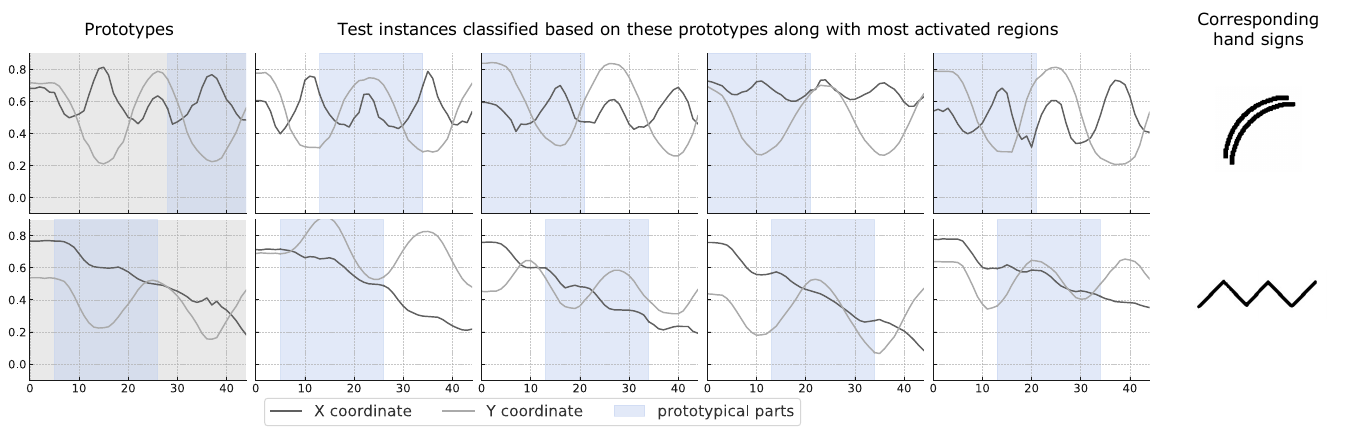}
    \caption{Two learned prototypes (left column) and, for each prototype, the four closest test instances (middle columns) with the most activated subsequences highlighted in blue (nearest latent patches by activation). The shown prototypes are those with the highest positive last-layer weight for their respective classes. The right column depicts the corresponding hand-sign sketches.}
    \label{fig:libras_protos_example}
\end{figure}

\begin{figure}
    \centering
    \includegraphics[width=\textwidth]{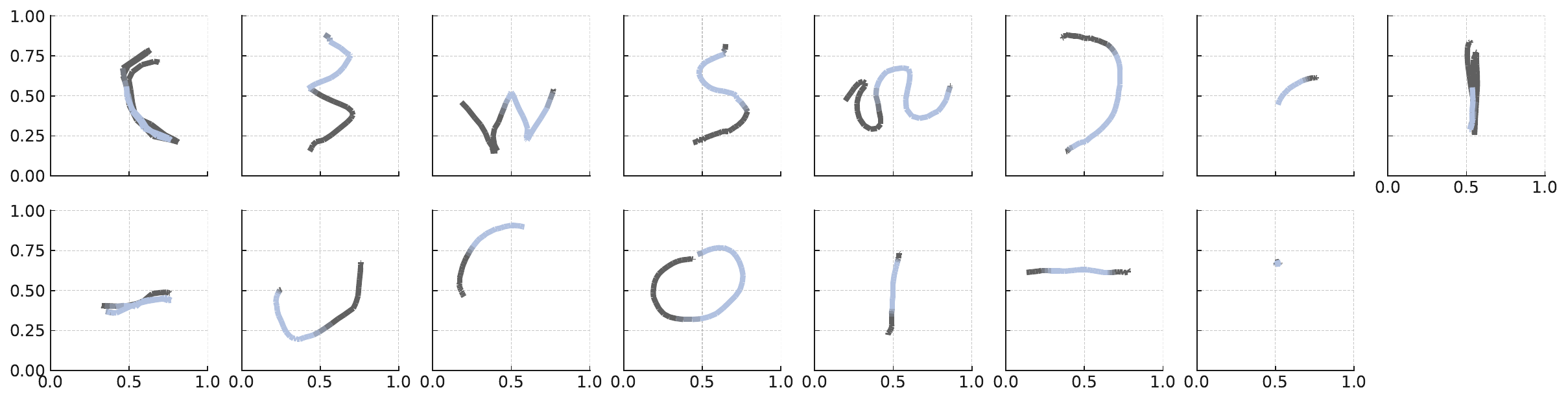}
    \caption{Prototypes for each class of the Libras dataset transformed and presented as 2D hand gestures with prototypical parts shown in blue.}
    \label{fig:libras_protos_all}
\end{figure}

\begin{figure}[p]
    \centering
    \includegraphics[width=0.98\textwidth]{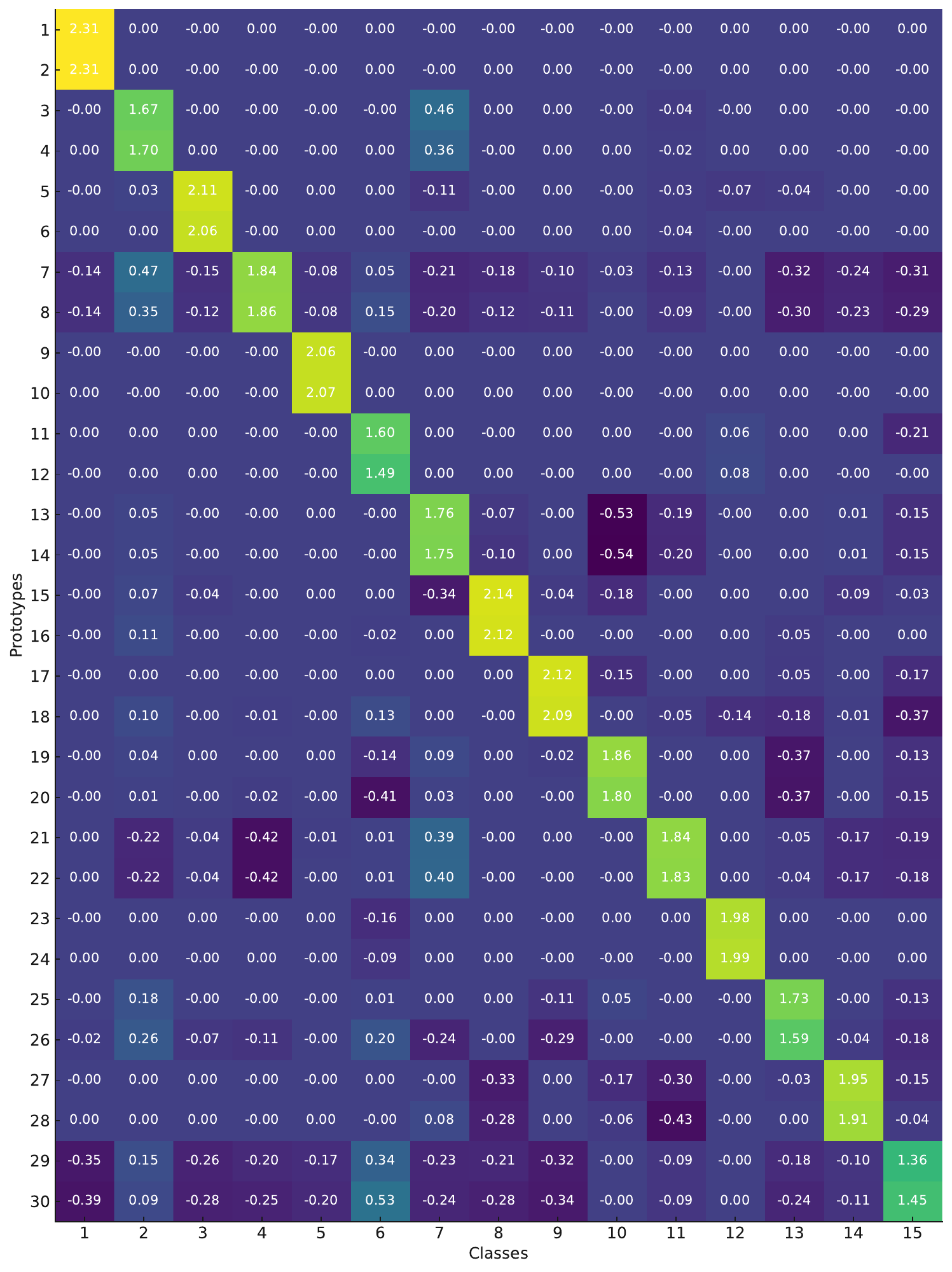}
    \caption{Last-layer weight matrix for the Libras experiment with two prototypes per class ($\lambda_{\text{last}}=3\times10^{-3}$).
    Columns correspond to classes (1-15) and rows to prototypes (1-30, ordered in class-wise pairs).
    Positive values (yellow) indicate "this resembles that" evidence; negative values (purple) indicate "does not resemble that".
    Biases are disabled in this layer. Diagonal block structure shows strong prototype-class associations.}
    \label{fig:libras-ll-weights}
\end{figure}

\textbf{Visualization and prototype selection.}
Figure~\ref{fig:libras_protos_example} presents two prototypes (each from a different class) together with four closest test instances for each (nearest latent patches by activation).
In the top row, the prototype captures a single arch segment that matches test trajectories, whereas the bottom-row prototype focuses on a short zig-zag (local reversals with $x$ cutting through $y$).
These prototypes were selected as those with the highest positive last-layer weight.
The corresponding last-layer weight matrix, highlighting prototype-class associations (positive "this resembles that" and negative "this does not resemble that" connections), is provided in Figure~\ref{fig:libras-ll-weights}.
Although the initialization sets positive weights for own-class connections and negative weights for other classes, there is no sign constraint during training.
Some off-diagonal connections become positive, indicating prototypes that provide supportive evidence for multiple classes, which can serve as an insight in expert analysis.

Figure~\ref{fig:libras_protos_all} shows one prototype per class (15 panels). For each class we display the prototype with the higher positive last-layer weight (out of the two class prototypes) and render its prototypical segment on the 2D hand trajectory. The remaining Libras prototypes are provided in the Appendix both as time series (Figure~\ref{fig:libras-all-prototypes-1d}) and as 2D trajectories (Figure~\ref{fig:libras-all-prototypes-2d}).

Inspection of the complete set of Libras prototypes (Appendix Figures~\ref{fig:libras-all-prototypes-1d} and \ref{fig:libras-all-prototypes-2d}) shows that, depending on the class, the two prototypes may: (i) cover essentially the same fragment of the same training instance, (ii) focus on different fragments within the same instance, (iii) capture distinct trajectories altogether. This variability, ranging from redundancy to complementarity, reappears across runs and can be informative for experts auditing the model (e.g., indicating stable motifs, localized patterns, or intra-class heterogeneity).

\textbf{Prototype-instance diagnostics.}
We attempt to quantify prototype-instance behavior by introducing a simple metric. 
\emph{Correct-Class Prototype Usage (CCPU)} measures how often the model’s most-activated prototype for an input belongs to the true class.
For class $y$:
$$
\mathrm{CCPU}(y) = \frac{1}{|\mathcal{D}^{\mathrm{test}}_y|}
\sum_{x\in\mathcal{D}^{\mathrm{test}}_y}
\mathbf{1}\!\Big[
\operatorname*{argmax}_{p\in\mathcal{P}} a_p(x)\ \in\ \mathcal{P}_y
\Big],
$$
where $a_p(x)$ is the prototype activation for $x$ (defined as similarity, Eq.~\ref{eq:similarity}), $\mathcal{D}^{\mathrm{test}}_y$ are test dataset instances belonging to class $y$, $\mathcal{P}$ is the full prototype set, $\mathcal{P}_c\subset\mathcal{P}$ are the prototypes of class $y$, and $\mathbf{1}$ is an indicator function.
On Libras, the median across 15 classes, averaged over five runs, is 90\%. Results are stable across the tested values of $\lambda_{\text{last}}$ (per-class values are presented in Table~\ref{tab:libras_per_class_diag}).

\begin{table}[tb]
    \centering
    \caption{Per-class \textit{Correct-Class Prototype Usage (CCPU)} for Libras dataset (two prototypes per class, $\lambda_{\text{last}}=3\times10^{-3}$).
    Each cell is the mean over 5 runs. CCPU is stable across $\lambda\in\{1\times10^{-3},3\times10^{-3},1\times10^{-2}\}$.}
    \label{tab:libras_per_class_diag}
    \setlength{\tabcolsep}{3pt}
    \begin{adjustbox}{width=\textwidth}
        \begin{tabular}{|l|r|r|r|r|r|r|r|r|r|r|r|r|r|r|r|}
            \hline
            Class & 1 & 2 & 3 & 4 & 5 & 6 & 7 & 8 & 9 & 10 & 11 & 12 & 13 & 14 & 15 \\
            \hline
            \hline
            CCPU & 0.88 & 0.08 & 0.77 & 1.00 & 0.75 & 0.85 & 0.40 & 0.98 & 1.00 & 0.83 & 0.92 & 0.88 & 0.87 & 0.97 & 1.00 \\
            \hline
        \end{tabular}
    \end{adjustbox}
\end{table}

\textbf{Simplicity-accuracy trade-off.}
Increasing the last-layer coefficient $\lambda_{\text{last}}$ simplifies the "this resembles that" logic by shrinking negative (off-diagonal) connections, at a cost of accuracy.
On Libras, accuracies are:
$\lambda_{\text{last}}{=}1{\times}10^{-3}$: 88.3\%,
$\lambda_{\text{last}}{=}3{\times}10^{-3}$: 85.6\%,
$\lambda_{\text{last}}{=}1{\times}10^{-2}$: 85.2\%.
(Values are lower than in Table~\ref{tab:accu_scores} because this analysis uses two prototypes per class for clearer visualization, whereas the main UEA experiments use ten.)

The weight matrix in Fig.~\ref{fig:libras-ll-weights} illustrates prototype-class associations for the middle setting. The corresponding matrices for the two remaining values of $\lambda_{\text{last}}$ are provided in Figures~\ref{fig:libras-ll-weights-low} and \ref{fig:libras-ll-weights-high} in the Appendix.
As $\lambda_{\text{last}}$ increases, negative off-diagonal connections are progressively pruned. At the low setting many non-zero negatives remain (the model frequently uses other-class prototypes to support evidence), at the middle setting only occasional off-diagonal entries persist, at the high setting nearly all negatives are zeroed, yielding an almost purely diagonal, "this resembles that"-only mapping. This simplification comes at an accuracy cost.

\subsection{Ablation Study}

To analyze the impact of our architectural choices, we conducted an ablation study comparing four ProtoTSNet variants:
\begin{itemize}
    \item full variant with grouping encoder and pretraining (GE/P),
    \item with grouping encoder without pretraining (GE/NP),
    \item with regular encoder and pretraining (RE/P),
    \item with regular encoder without pretraining (RE/NP).
\end{itemize}

All variants used the hyperparameters optimized in previous experiments. The~training process consisted of four cycles of (joint epochs $\rightarrow$ prototype projection $\rightarrow$ last layer optimization), totaling approximately 360 epochs. When applied, encoder pretraining comprised 50 epochs.

Table \ref{tab:ablation_results} included in the Appendix presents the classification accuracy across all datasets. The full variant achieves the best performance among the tested variants with an average rank of 1.90 and 13 wins/ties, followed by the regular encoder without pretraining variant with an average rank of 2.33 and 11 wins/ties.

An interesting observation emerges regarding the impact of pretraining across encoder architectures. Pretraining significantly improves performance with the grouping encoder, but shows a small impact with the regular encoder. This discrepancy may be explained by the pretraining objective: as part of an autoencoder, the encoder learns to represent all features with equal significance. While this might not be optimal for classification, where feature importance varies, the grouping encoder can later selectively disable certain feature groups during ProtoTSNet training thanks to the~$1 \times 1$ convolution layer following the encoder.

While the introduction of the grouping encoder initially degrades performance (the~non-pretrained grouped encoder ranks last in both average rank and wins/ties), the~addition of pretraining not only recovers, but surpasses the regular encoder's performance. The performance varies across datasets: the full variant excels on datasets like Epilepsy and LSST, while the regular encoder performs better on EigenWorms and Handwriting. Some datasets, such as ArticularyWordRecognition, Heartbeat, and JapaneseVowels, show comparable performance across variants.

\begin{figure}
    \centering
    \includegraphics[width=0.75\textwidth]{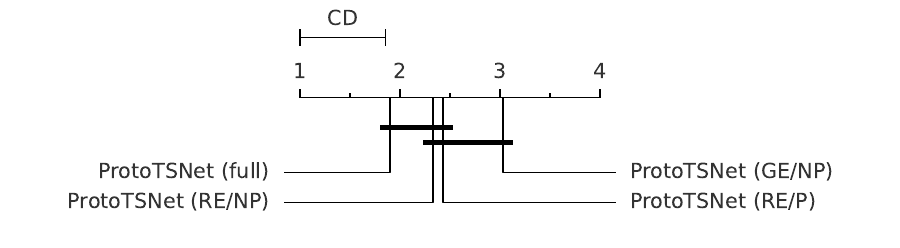}
    \caption{Critical difference diagram for the four ProtoTSNet variants evaluated on the UEA multivariate datasets ($\alpha = 0.05$). The full variant (GE/P) shows statistically significant improvement over the grouped encoder without pretraining variant (GE/NP), while other comparisons show no significant differences. Abbreviations used: \textit{GE} - modified encoder with grouping, \textit{RE} - regular encoder without grouping, \textit{P} - encoder pretraining, \textit{NP} - no encoder pretraining. The full variant combines grouping encoder with pretraining.}
    \label{fig:ablation_only_cd_diag}
\end{figure}

The critical difference analysis using Friedman test followed by a Nemenyi pairwise post-hoc test for multiple comparisons of mean rank sums~\citep{demsar2006statistical} presented in Figure~\ref{fig:ablation_only_cd_diag} compares the average ranks across the four ProtoTSNet variants. The~full variant (GE/P) achieves the best average rank of 1.90 and shows a statistically significant improvement over the grouped encoder without pretraining variant (GE/NP).

\subsubsection{Parameter Sensitivity Analysis}

Our architecture incorporates two key parameters that influence the model behavior:

\begin{itemize}
    \item prototype length: the length of individual prototypes expressed as a percentage of input time series length,
    \item reception: the percentage of input features passed to each encoder group, controlling the sparsity of grouping encoder masks.
\end{itemize}

\begin{figure}[tb]
    \centering
    \subfloat{
        \centering
        \begin{minipage}[c][\height][c]{0.48\textwidth}
            \centering
            ArticularyWordRecognition
            \includegraphics[width=\textwidth]{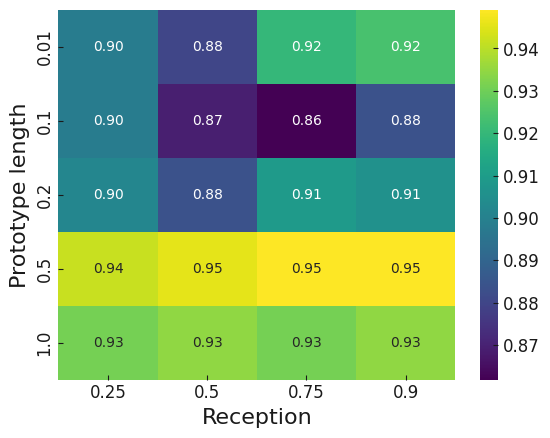}
        \end{minipage}
    }
    \subfloat{
        \centering
        
        \begin{minipage}[c][\height][c]{0.48\textwidth}
            \centering
            EigenWorms
            \includegraphics[width=\textwidth]{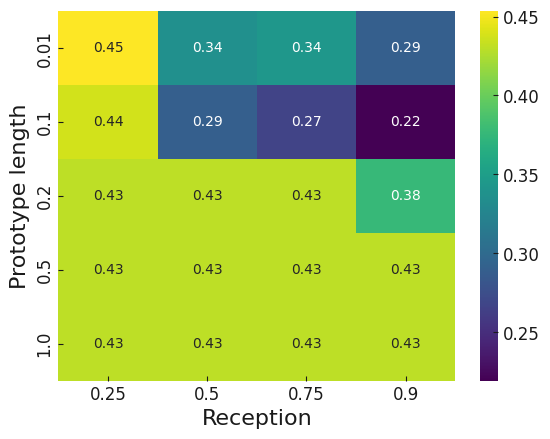}
        \end{minipage}
    }

    \subfloat{
        \centering
        \begin{minipage}[c][\height][c]{0.48\textwidth}
            \centering
            ERing
            \includegraphics[width=\textwidth]{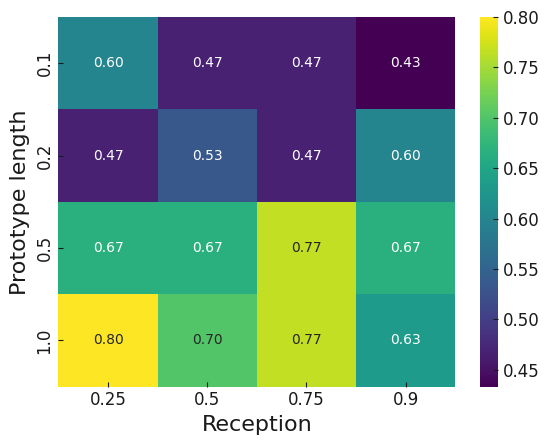}
        \end{minipage}
    }
    \subfloat{
        \centering
        
        \begin{minipage}[c][\height][c]{0.48\textwidth}
            \centering
            SelfRegulationSCP1
            \includegraphics[width=\textwidth]{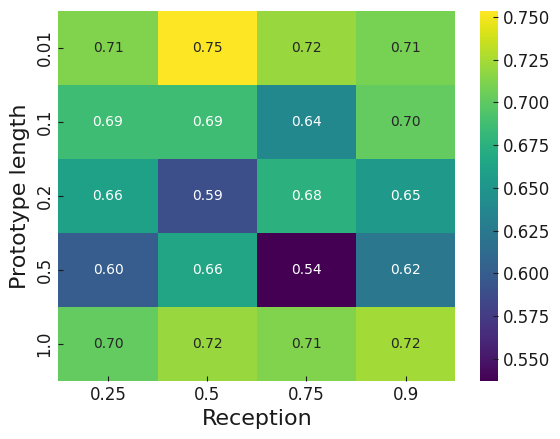}
        \end{minipage}
    }
    \caption{Parameter sensitivity analysis showing classification accuracy across prototype length and reception parameter values for four selected datasets. Heatmaps demonstrate varying sensitivity patterns across different dataset characteristics. Complete analysis with feature importance bar charts for all datasets is provided in Appendix Figure~\ref{fig:ablation_params}.}
    \label{fig:param_sensitivity_main}
\end{figure}

Figure~\ref{fig:param_sensitivity_main} presents parameter sensitivity analysis across four representative datasets using heatmaps of accuracy across the 2D parameter grid. Additional heatmaps and corresponding bar charts of feature importance for all datasets are provided in Appendix Figure~\ref{fig:ablation_params}. All values represent averages from five cross-validation runs.

The impact of prototype length on classification accuracy varies significantly across datasets. For datasets like ERing and SelfRegulationSCP1 (Figure~\ref{fig:param_sensitivity_main}), prototype length substantially influences performance, reflecting the varying lengths of class-characteristic patterns. Some datasets, such as EigenWorms, show good performance with longer prototypes as well, suggesting their ability to capture shorter characteristic patterns. Others, like CharacterTrajectories and SelfRegulationSCP1, exhibit multiple prototype lengths yielding comparable accuracy. This parameter can be effectively tuned using domain expertise, as it corresponds to the expected length of class-distinctive temporal patterns.

The reception parameter's influence on accuracy is generally more subtle and dataset-dependent. While some datasets (ArticularyWordRecognition, EthanolConcentration) show minimal sensitivity to this parameter, others (LSST, EigenWorms) demonstrate more pronounced effects. However, reception strongly influences feature importance calculations, which aligns with its role in controlling feature filtering within the grouping encoder. Higher reception values can lead to overestimated feature importance scores due to increased feature mixing. For practical applications, we recommend initializing this parameter with a high value and gradually reducing it until accuracy begins to degrade (accuracy may initially improve with reduction).

\section{Summary}
\label{summary}

In this paper, we presented ProtoTSNet, a novel approach for interpretable multivariate time series classification that provides ante hoc explanations through prototypical parts. Our method addresses a significant gap in the field of explainable time series analysis, where ante hoc approaches for multivariate data are particularly scarce. The~key advantages of our approach include the ability to identify significant temporal patterns through prototypes, calculate feature importance, and provide explanations that are accessible to domain experts without requiring a deep understanding of the~underlying algorithms.

Our evaluation on the UEA multivariate time series archive demonstrates that ProtoTSNet achieves the best performance among ante-hoc explainable methods, followed by the shapelet-based approach, which despite its interpretability relies on rotation forests that may complicate direct explanation understanding. While some methods such as ROCKET (non-explainable) and LITEMVTime (post-hoc explainable) achieve higher average ranks, ProtoTSNet maintains competitive performance while providing the advantage of direct ante-hoc interpretability. Through ablation studies, we validated our architectural choices and provided insights into parameter selection through sensitivity analysis. Statistical analysis using critical difference diagrams confirms the~significance of our results.

The prototype-based nature of our explanations offers direct interpretability through the "this resembles that" principle, providing clearer explanations than methods that rely on ensemble techniques or complex feature transformations. This opens up interesting possibilities for integration with symbolic approaches, potentially bridging the gap between neural and symbolic representations in time series analysis. Our future work will focus on domain-specific applications and user studies to evaluate the intuitiveness of the model's explanations, enhancing our understanding of the model's capabilities and limitations in practical contexts.

\section*{Acknowledgments}
This paper is part of a project that has received funding from the European Union's Horizon Europe Research and Innovation Programme, under Grant Agreement number 101120406. The paper reflects only the authors' view and the EC is not responsible for any use that may be made of the information it contains.
The research has been supported by a grant from the Priority Research Area (DigiWorld) under the Strategic Programme Excellence Initiative at Jagiellonian University.

\bibliography{bibliography}

\newpage

\begin{appendices}
\renewcommand{\theHfigure}{A.\arabic{figure}}
\renewcommand{\theHtable}{A.\arabic{table}}

\section{Datasets quantitative description and results}\label{secA1}

\subsection{Datasets and per-dataset hyperparameters}

\begin{table}[ht!]
    \caption{UEA multivariate datasets characteristics together with the hyperparameters found through a 5-fold cross-validation. \textit{Train} stands for the training set size, \textit{test} for the test set size, $r$ stands for reception of the encoder and $L$ stands for prototype length as a fraction of time series length.}
    \label{tab:datasets_characteristics}
    \centering
    \setlength{\tabcolsep}{5pt}
    \begin{tabular}{|l|r|r|r|r|r|r|r|}
        \hline
        Dataset & Train & Test & Dims & Len & Classes & $r$ & $L$ \\
        \hline
        \hline
        ArticularyWordRecognition & 275 & 300 & 9 & 144 & 25 & 0.75 & 0.5 \\
        AtrialFibrillation & 15 & 15 & 2 & 640 & 3 & 0.5 & 0.2 \\
        BasicMotions & 40 & 40 & 6 & 100 & 4 & 0.25 & 0.2 \\
        CharacterTrajectories & 1422 & 1436 & 3 & 182 & 20 & 0.9 & 1 \\
        Cricket & 108 & 72 & 6 & 1197 & 12 & 0.75 & 0.1 \\
        DuckDuckGeese & 50 & 50 & 1345 & 270 & 5 & 0.5 & 0.01 \\
        ERing & 30 & 270 & 4 & 65 & 6 & 0.25 & 1 \\
        EigenWorms & 128 & 131 & 6 & 17984 & 5 & 0.25 & 0.01 \\
        Epilepsy & 137 & 138 & 3 & 206 & 4 & 0.5 & 1 \\
        EthanolConcentration & 261 & 263 & 3 & 1751 & 4 & 0.25 & 0.01 \\
        FaceDetection & 5890 & 3524 & 144 & 62 & 2 & 0.9 & 1 \\
        FingerMovements & 316 & 100 & 28 & 50 & 2 & 0.5 & 1 \\
        HandMovementDirection & 160 & 74 & 10 & 400 & 4 & 0.5 & 1 \\
        Handwriting & 150 & 850 & 3 & 152 & 26 & 0.75 & 0.5 \\
        Heartbeat & 204 & 205 & 61 & 405 & 2 & 0.5 & 1 \\
        InsectWingbeat & 30000 & 20000 & 200 & 30 & 10 & 0.25 & 0.1 \\
        JapaneseVowels & 270 & 370 & 12 & 29 & 9 & 0.5 & 1 \\
        LSST & 2459 & 2466 & 6 & 36 & 14 & 0.75 & 0.1 \\
        Libras & 180 & 180 & 2 & 45 & 15 & 0.9 & 0.5 \\
        MotorImagery & 278 & 100 & 64 & 3000 & 2 & 0.5 & 0.01 \\
        NATOPS & 180 & 180 & 24 & 51 & 6 & 0.5 & 0.5 \\
        PEMS-SF & 267 & 173 & 963 & 144 & 7 & 0.75 & 1 \\
        PenDigits & 7494 & 3498 & 2 & 8 & 10 & 0.75 & 0.5 \\
        PhonemeSpectra & 3315 & 3353 & 11 & 217 & 39 & 0.75 & 0.1 \\
        RacketSports & 151 & 152 & 6 & 30 & 4 & 0.9 & 0.1 \\
        SelfRegulationSCP1 & 268 & 293 & 6 & 896 & 2 & 0.5 & 0.01 \\
        SelfRegulationSCP2 & 200 & 180 & 7 & 1152 & 2 & 0.5 & 0.2 \\
        SpokenArabicDigits & 6599 & 2199 & 13 & 93 & 10 & 0.75 & 1 \\
        StandWalkJump & 12 & 15 & 4 & 2500 & 3 & 0.75 & 0.5 \\
        UWaveGestureLibrary & 120 & 320 & 3 & 315 & 8 & 0.75 & 0.5 \\
        \hline
    \end{tabular}
\end{table}

\newpage

\subsection{Common optimization hyperparameters}
\label{secA:common-params}

This subsection lists the optimization and last-layer settings that were
identical across all datasets. Parameters that varied are described in Section~\ref{sec:uea_datasets}.

\paragraph{Optimizer and learning rates.}
Adam optimizer. During warmup we used a constant learning rate of $3\times10^{-3}$. In the joint-training and last-layer phases we used
PyTorch \texttt{CyclicLR} with learning rate swept from $1\times10^{-4}$ to
$1\times10^{-2}$ (triangular schedule peaking near the middle of each cycle).
No weight decay. No momentum scheduling.

\paragraph{Epoch schedule.}
Fixed schedule for all datasets:
(i) \emph{Pretraining} (autoencoder): 50 epochs (ablation variants without pretraining skip this);
(ii) \emph{Warmup}: 50 epochs at constant LR (only for pretrained variants, otherwise these were joint epochs);
(iii) Four cycles of \emph{joint training} $\rightarrow$ \emph{prototype projection}
\(\rightarrow\) \emph{last-layer optimization}. The per-cycle composition was the same
for every dataset; only $r$ and $L$ varied (Table~\ref{tab:datasets_characteristics}).

\paragraph{Regularization.}
Prototype clustering $\lambda_{\text{clst}}=0.08$;
prototype separation $\lambda_{\text{sep}}=-0.008$;
L1 on the last layer $\lambda_{\text{last}}=1\times10^{-3}$;
L1 on the feature-mixing layer $\lambda_{\text{conv}}=1\times10^{-3}$.
All coefficients were constant (no annealing or thresholding schedules).

\paragraph{Batch size rule.}
Default batch size $=32$. If this exceeded $25\%$ of the dataset’s training set size,
the largest power of two smaller than $25\%$ of that size was used instead,
ensuring at least four batches per epoch.

\paragraph{Prototype configuration and layers initialization.}
The number of prototypes was fixed to 10 per class in all experiments. All network parameters in layers preceding the final classification layer, including prototypes, were initialized randomly using Kaiming normal initialization. The final classification layer weights were initialized to 1 for positive prototype-class connections and -0.5 for negative ones, with the bias terms disabled in both this layer and the $1 \times 1$ feature-mixing convolution layer.

\newpage

\subsection{Results and ablations}

\begin{table}[ht]
    \caption{Accuracy scores achieved by the methods we tested on the UEA multivariate datasets. Best results in \textbf{bold}. Abbreviations for methods: \textit{PTS} - ProtoTSNet, \textit{PET} - PETSC, \textit{Sha} - Shaplet-based, \textit{MTC} - MTEX-CNN, \textit{Lite} - LITEMVTime, \textit{Tap} - TapNet, \textit{ROC} - ROCKET.}
    \label{tab:accu_scores}
    \centering
    \begin{adjustbox}{width=\textwidth}
        \begin{tabular}{|l|c|c|c|c|c|c|c|c|}
            \hline
            \multirow{2}{*}{Dataset} & \multicolumn{3}{c|}{ante hoc} & \multicolumn{3}{c|}{post hoc} & \multicolumn{2}{c|}{non-explainable} \\
            & PTS & PET & Sha & XCM & MTC & Lite & Tap & ROC \\
            \hline
            \hline
            ArticularyWordRecognition & 96.7 & \textbf{99.3} & 97.0 & 72.9 & 94.2 & 98.7 & 98.3 & \textbf{99.3} \\
            AtrialFibrillation & \textbf{33.3} & 30.7 & 32.0 & 30.7 & \textbf{33.3} & 21.3 & 25.3 & 6.7 \\
            BasicMotions & \textbf{100.0} & 94.5 & 94.5 & 95.5 & 97.0 & \textbf{100.0} & \textbf{100.0} & \textbf{100.0} \\
            CharacterTrajectories & 98.1 & 93.3 & 97.6 & 95.7 & 95.8 & 99.6 & \textbf{99.7} & N/A \\
            Cricket & 94.4 & 99.4 & 97.2 & 71.7 & 89.4 & 98.6 & 97.8 & \textbf{100.0} \\
            DuckDuckGeese & \textbf{68.8} & 42.0 & 43.6 & 31.6 & 51.2 & N/A & 48.8 & 52.0 \\
            EigenWorms & 38.9 & N/A & N/A & 28.2 & 42.0 & 68.9 & 47.8 & \textbf{90.8} \\
            Epilepsy & 96.8 & 98.4 & 98.8 & 89.7 & 92.8 & \textbf{99.6} & 97.0 & 99.3 \\
            ERing & 82.2 & 88.3 & 92.5 & 20.6 & 84.5 & 93.1 & 82.0 & \textbf{98.5} \\
            EthanolConcentration & 29.4 & 50.6 & \textbf{75.7} & 28.8 & 28.5 & 26.3 & 28.0 & 44.5 \\
            FaceDetection & 63.3 & 56.2 & N/A & 55.5 & 50.0 & \textbf{65.0} & 51.3 & 64.4 \\
            FingerMovements & 53.8 & 52.6 & 55.8 & 51.4 & 49.0 & \textbf{62.8} & 49.6 & 54.0 \\
            HandMovementDirection & \textbf{53.0} & 25.9 & 49.5 & 32.4 & 36.5 & 40.3 & 35.9 & 51.4 \\
            Handwriting & 20.2 & 40.2 & 40.6 & 17.6 & 21.5 & \textbf{64.1} & 53.0 & 59.5 \\
            Heartbeat & 72.0 & 67.4 & 72.0 & 71.8 & 66.3 & 67.4 & 73.6 & \textbf{74.6} \\
            InsectWingbeat & 64.9 & N/A & N/A & 61.2 & 69.5 & \textbf{73.5} & 47.2 & N/A \\
            JapaneseVowels & 97.2 & N/A & 87.8 & 94.1 & 93.5 & \textbf{98.9} & 98.1 & N/A \\
            Libras & 90.1 & 81.6 & 87.9 & 59.1 & 64.6 & \textbf{92.1} & 85.4 & 90.0 \\
            LSST & 60.2 & 54.1 & \textbf{64.1} & 56.7 & 40.8 & 31.3 & 46.8 & 63.7 \\
            MotorImagery & 52.8 & N/A & 51.2 & 54.0 & 50.0 & 51.0 & 48.6 & \textbf{58.0} \\
            NATOPS & 94.8 & 83.0 & 85.7 & 83.4 & 89.0 & 95.9 & \textbf{96.4} & 88.9 \\
            PEMS-SF & 84.9 & 91.3 & \textbf{95.6} & 69.4 & 49.5 & 71.9 & 76.6 & 83.2 \\
            PenDigits & 97.0 & 86.7 & N/A & 95.7 & 89.4 & \textbf{98.6} & 97.1 & 98.3 \\
            PhonemeSpectra & 23.9 & 24.3 & N/A & 10.5 & 6.2 & \textbf{31.1} & 9.2 & 27.6 \\
            RacketSports & 83.4 & 83.2 & 83.0 & 66.3 & 69.5 & 86.7 & 84.3 & \textbf{89.5} \\
            SelfRegulationSCP1 & 76.0 & 69.7 & 83.7 & 68.3 & 67.6 & 80.2 & 65.8 & \textbf{84.3} \\
            SelfRegulationSCP2 & 50.9 & 49.4 & 52.0 & 52.6 & 50.0 & 53.4 & 54.3 & \textbf{57.2} \\
            SpokenArabicDigits & 94.0 & N/A & 67.0 & 97.6 & 97.2 & \textbf{99.6} & 98.4 & N/A \\
            StandWalkJump & 34.7 & 37.8 & 50.7 & 24.0 & 42.7 & 44.0 & 26.7 & \textbf{53.3} \\
            UWaveGestureLibrary & 84.1 & 88.8 & 88.1 & 81.4 & 82.6 & 91.4 & 91.9 & \textbf{94.1} \\
            \hline
            \hline
            Avg. Rank & 3.90 & 5.30 & 4.43 & 6.00 & 5.73 & 3.03 & 4.33 & \textbf{2.73} \\
            Wins/Ties & 3 & 0 & 3 & 0 & 1 & 11 & 3 & \textbf{12} \\
            \hline
        \end{tabular}
    \end{adjustbox}
\end{table}

\begin{table}[ht]
    \caption{Classification accuracy scores for four ProtoTSNet variants on UEA multivariate datasets. Best results are shown in \textbf{bold}. Variant naming convention: \textit{GE} - grouping encoder, \textit{RE} - regular encoder, \textit{P} - with pretraining, \textit{NP} - without pretraining. The full variant (GE/P) combines grouping encoder with pretraining.}
    \label{tab:ablation_results}
    \centering
    \begin{tabular}{|l|c|c|c|c|}
        \hline
        Dataset & GE/P & GE/NP & RE/P & RE/NP \\
        \hline
        \hline
        ArticularyWordRecognition & 96.7 & 96.5 & 96.5 & \textbf{96.8} \\
        AtrialFibrillation & 33.3 & \textbf{36.0} & 26.7 & 33.3 \\
        BasicMotions & \textbf{100.0} & \textbf{100.0} & \textbf{100.0} & \textbf{100.0} \\
        CharacterTrajectories & \textbf{98.1} & 97.9 & 97.9 & 97.8 \\
        Cricket & 94.4 & 91.4 & \textbf{94.7} & 92.5 \\
        DuckDuckGeese & \textbf{68.8} & 62.8 & 68.0 & 64.4 \\
        EigenWorms & 38.9 & 32.2 & 28.9 & \textbf{57.7} \\
        Epilepsy & \textbf{96.8} & 87.7 & 88.1 & 80.0 \\
        ERing & \textbf{82.2} & 80.3 & 78.8 & 82.0 \\
        EthanolConcentration & \textbf{29.4} & 28.7 & \textbf{29.4} & 28.6 \\
        FaceDetection & \textbf{63.3} & 59.9 & 61.1 & 60.9 \\
        FingerMovements & \textbf{53.8} & 52.0 & 51.0 & 52.0 \\
        HandMovementDirection & 53.0 & 50.8 & \textbf{53.8} & 42.2 \\
        Handwriting & 20.2 & 19.4 & 20.8 & \textbf{25.0} \\
        Heartbeat & 72.0 & \textbf{72.2} & 71.7 & 71.8 \\
        InsectWingbeat & 64.9 & 63.1 & \textbf{65.4} & 64.8 \\
        JapaneseVowels & 97.2 & 96.8 & \textbf{97.6} & 97.2 \\
        Libras & \textbf{90.1} & 88.6 & 90.0 & 87.6 \\
        LSST & \textbf{60.2} & 53.1 & 55.9 & 56.9 \\
        MotorImagery & 52.8 & 50.6 & 51.6 & \textbf{53.8} \\
        NATOPS & 94.8 & 94.7 & 94.9 & \textbf{96.4} \\
        PEMS-SF & 84.9 & \textbf{86.9} & 86.8 & 85.8 \\
        PenDigits & 97.0 & 96.7 & \textbf{98.6} & \textbf{98.6} \\
        PhonemeSpectra & 23.9 & 21.7 & 23.7 & \textbf{24.0} \\
        RacketSports & 83.4 & 83.8 & 85.5 & \textbf{85.8} \\
        SelfRegulationSCP1 & \textbf{76.0} & 73.8 & 75.0 & 72.5 \\
        SelfRegulationSCP2 & 50.9 & 50.4 & 48.2 & \textbf{52.6} \\
        SpokenArabicDigits & 94.0 & 96.5 & 96.2 & \textbf{97.3} \\
        StandWalkJump & \textbf{34.7} & 33.3 & 28.0 & \textbf{34.7} \\
        UWaveGestureLibrary & \textbf{84.1} & 82.8 & 83.2 & 80.7 \\
        \hline
        \hline
        Avg. Rank & \textbf{1.90} & 3.03 & 2.43 & 2.33 \\
        Wins/Ties & \textbf{13} & 4 & 6 & 11 \\
        \hline
    \end{tabular}
\end{table}

\begin{figure}[t]
    \centering

    ArticularyWordRecognition
    \subfloat{
        \centering
        \begin{minipage}[c][\height][c]{0.38\textwidth}
            \includegraphics[width=\textwidth]{assets/ablation/heatmaps/ArticularyWordRecognition.png}
        \end{minipage}
    }
    \subfloat{
        \centering
        \begin{minipage}[c][\height][c]{0.58\textwidth}
            \includegraphics[width=\textwidth]{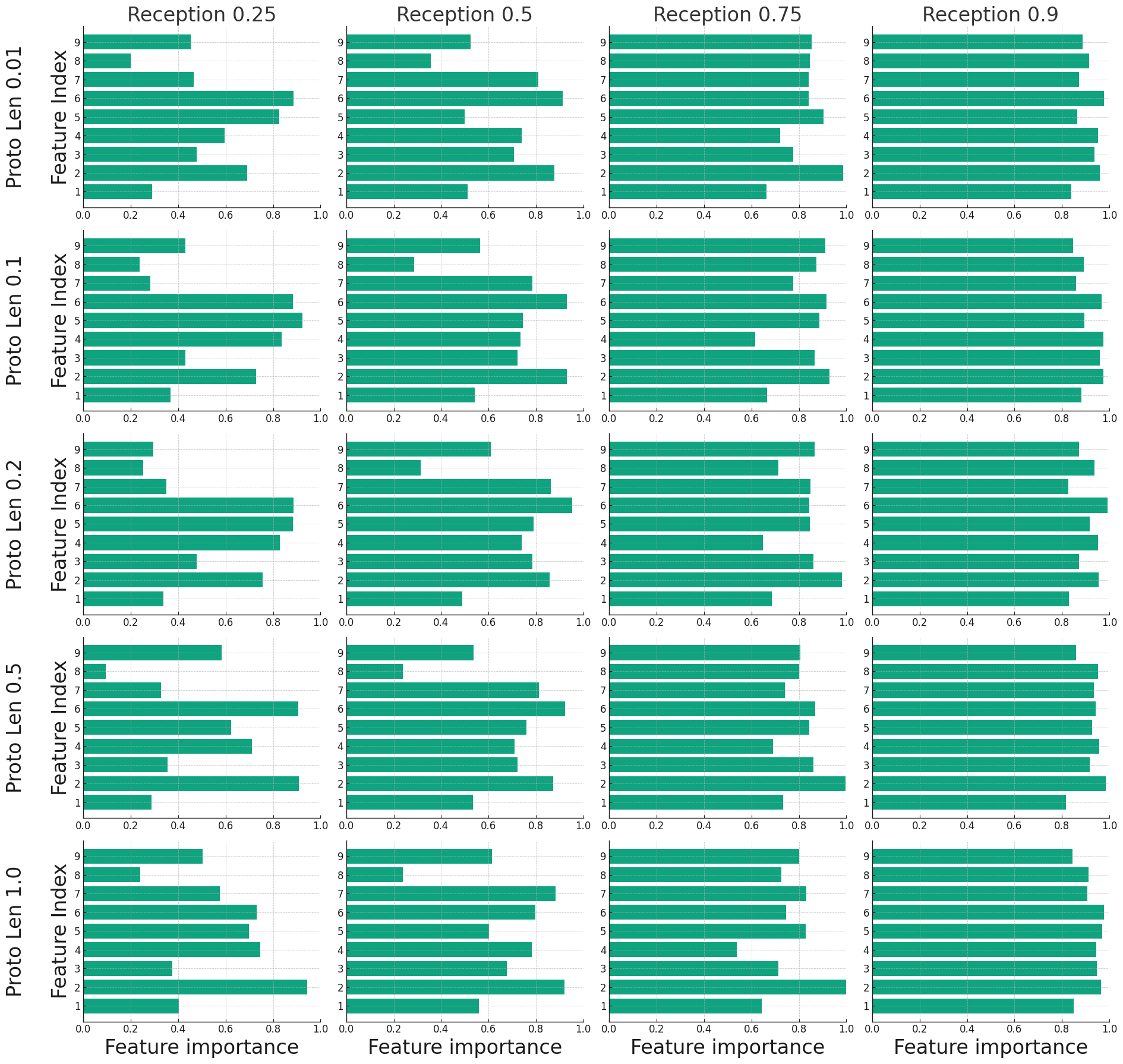}
        \end{minipage}
    }

    \bigskip
    CharacterTrajectories
    \subfloat{
        \centering
        \begin{minipage}[c][\height][c]{0.38\textwidth}
            \includegraphics[width=\textwidth]{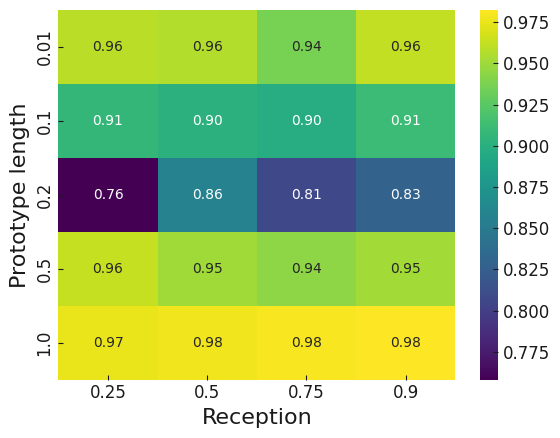}
        \end{minipage}
    }
    \subfloat{
        \centering
        \begin{minipage}[c][\height][c]{0.58\textwidth}
            \includegraphics[width=\textwidth]{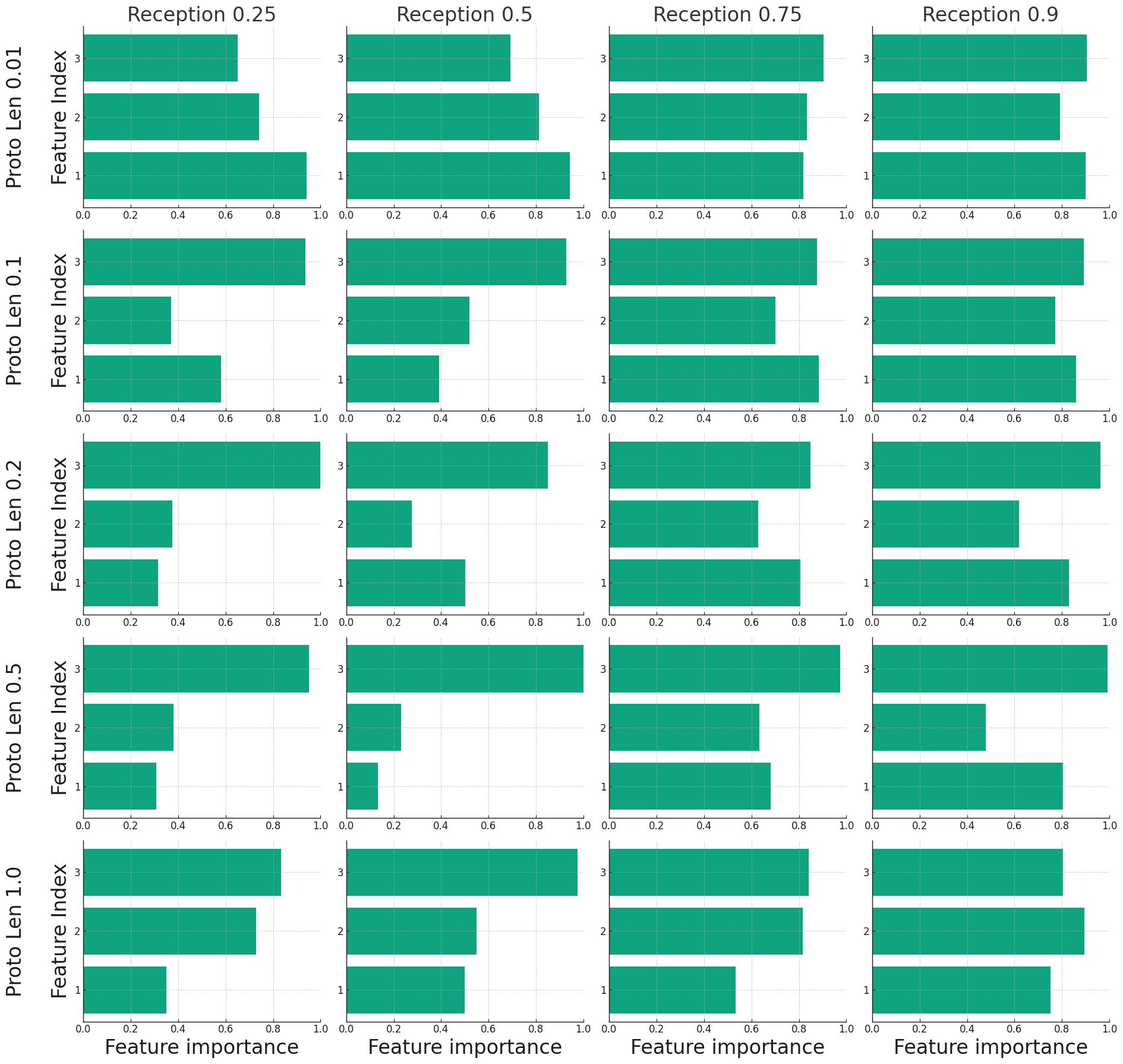}
        \end{minipage}
    }

    \caption{Parameter sensitivity analysis for selected datasets, examining prototype length and reception parameters. Left: Heatmaps showing average classification accuracy across the parameter grid. Right: Bar charts showing average feature importance scores across the same parameter combinations. The parameter grid explores five prototype lengths (as fractions of input time series length) and four reception values (controlling feature sparsity in the grouping encoder). All values are averaged over five cross-validation runs.}
    \label{fig:ablation_params}
\end{figure}
\begin{figure}[t]\ContinuedFloat
    \centering

    Cricket
    \subfloat{
        \centering
        \begin{minipage}[c][\height][c]{0.38\textwidth}
            \includegraphics[width=\textwidth]{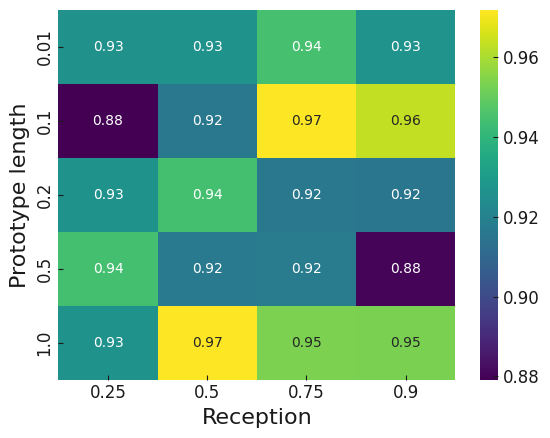}
        \end{minipage}
    }
    \subfloat{
        \centering
        \begin{minipage}[c][\height][c]{0.58\textwidth}
            \includegraphics[width=\textwidth]{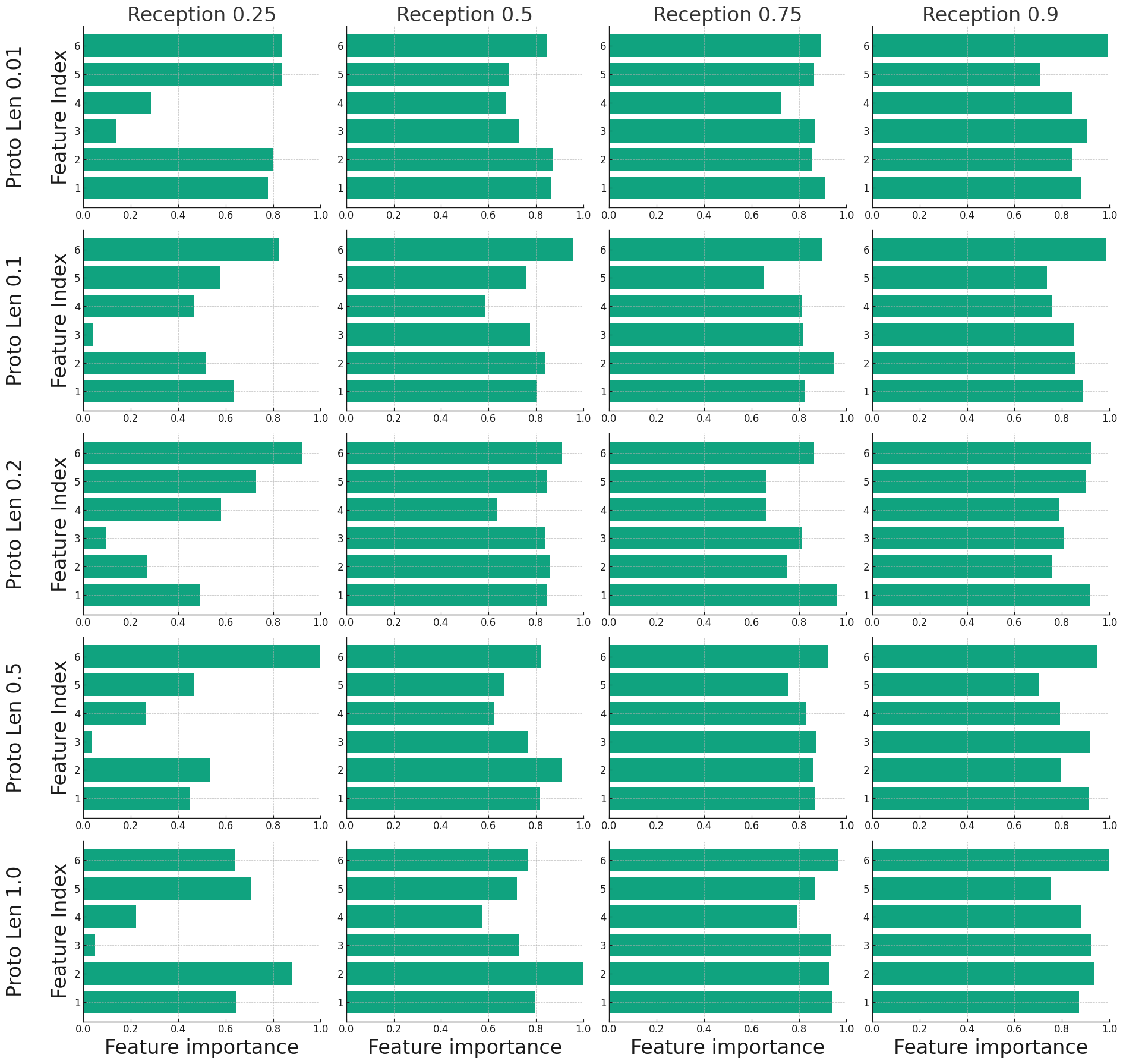}
        \end{minipage}
    }

    \bigskip
    EigenWorms
    \subfloat{
        \centering
        \begin{minipage}[c][\height][c]{0.38\textwidth}
            \includegraphics[width=\textwidth]{assets/ablation/heatmaps/EigenWorms.png}
        \end{minipage}
    }
    \subfloat{
        \centering
        \begin{minipage}[c][\height][c]{0.58\textwidth}
            \includegraphics[width=\textwidth]{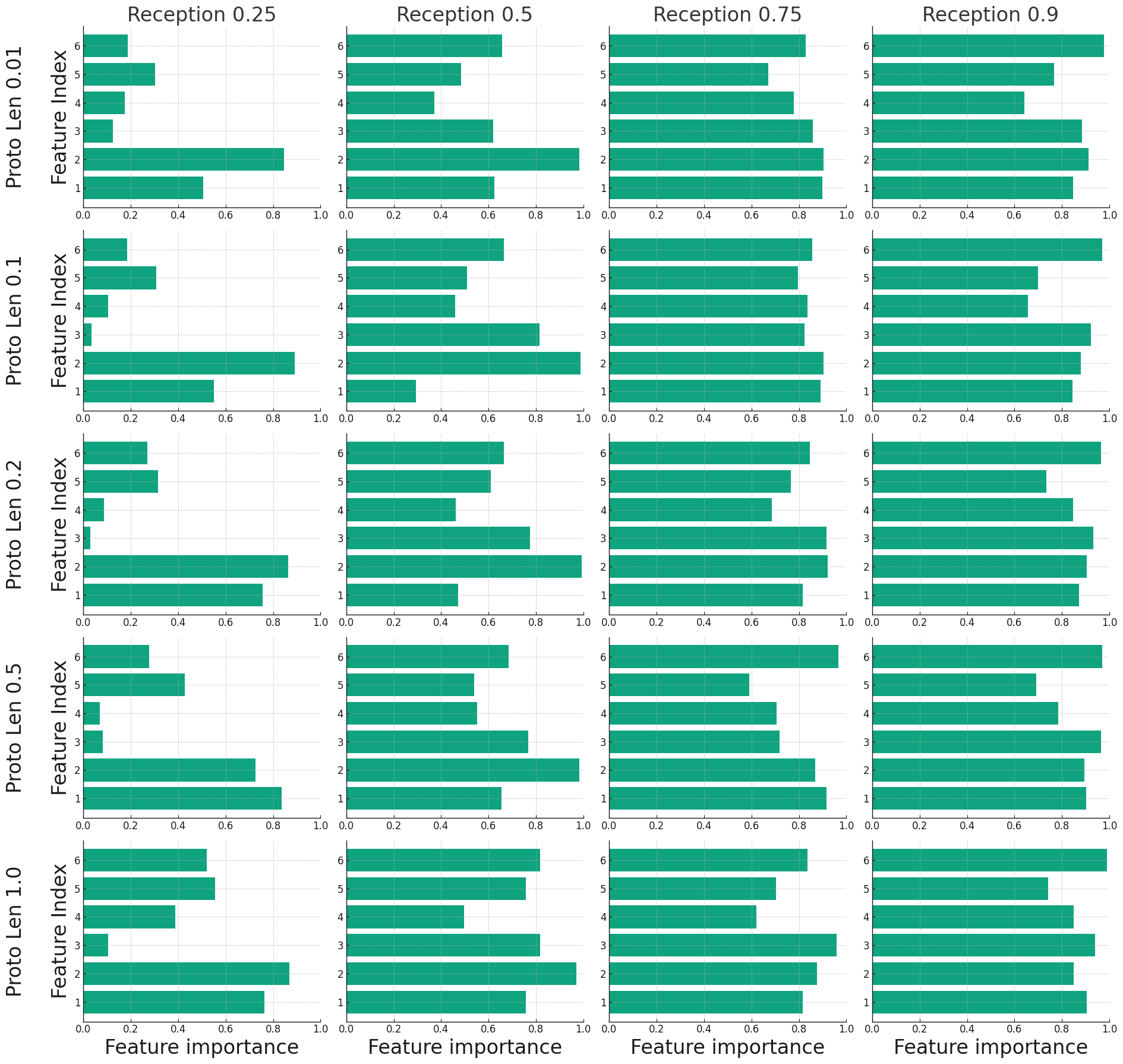}
        \end{minipage}
    }

    \caption{(contd)}
\end{figure}
\begin{figure}[t]\ContinuedFloat
    \centering

    ERing
    \subfloat{
        \centering
        \begin{minipage}[c][\height][c]{0.38\textwidth}
            \includegraphics[width=\textwidth]{assets/ablation/heatmaps/ERing.png}
        \end{minipage}
    }
    \subfloat{
        \centering
        \begin{minipage}[c][\height][c]{0.58\textwidth}
            \includegraphics[width=\textwidth]{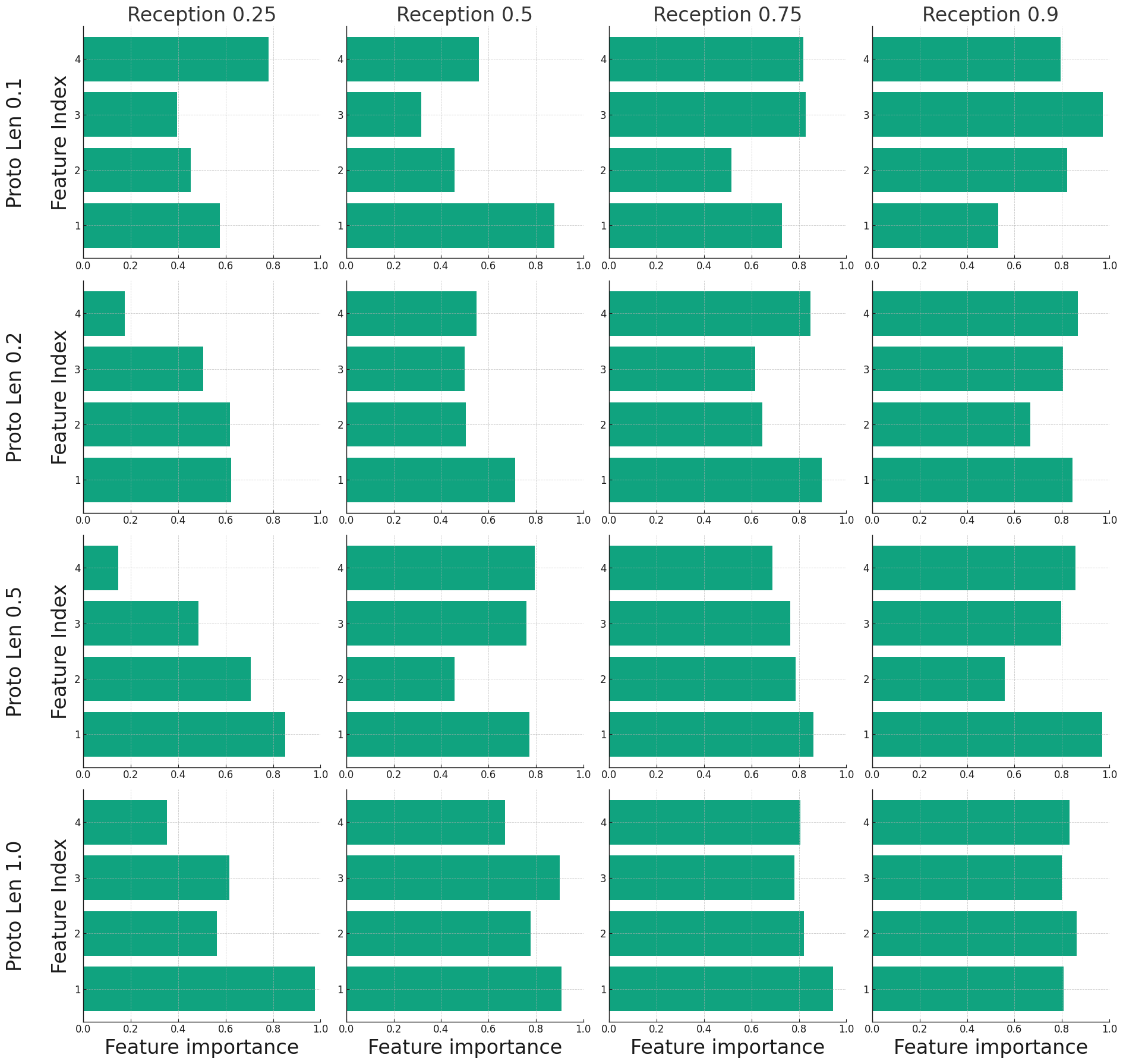}
        \end{minipage}
    }

    \bigskip
    EthanolConcentration
    \subfloat{
        \centering
        \begin{minipage}[c][\height][c]{0.38\textwidth}
            \includegraphics[width=\textwidth]{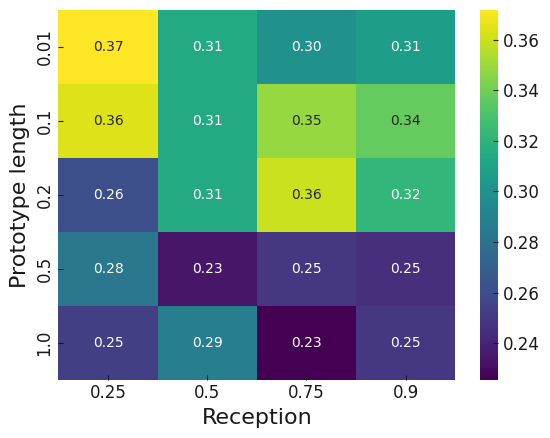}
        \end{minipage}
    }
    \subfloat{
        \centering
        \begin{minipage}[c][\height][c]{0.58\textwidth}
            \includegraphics[width=\textwidth]{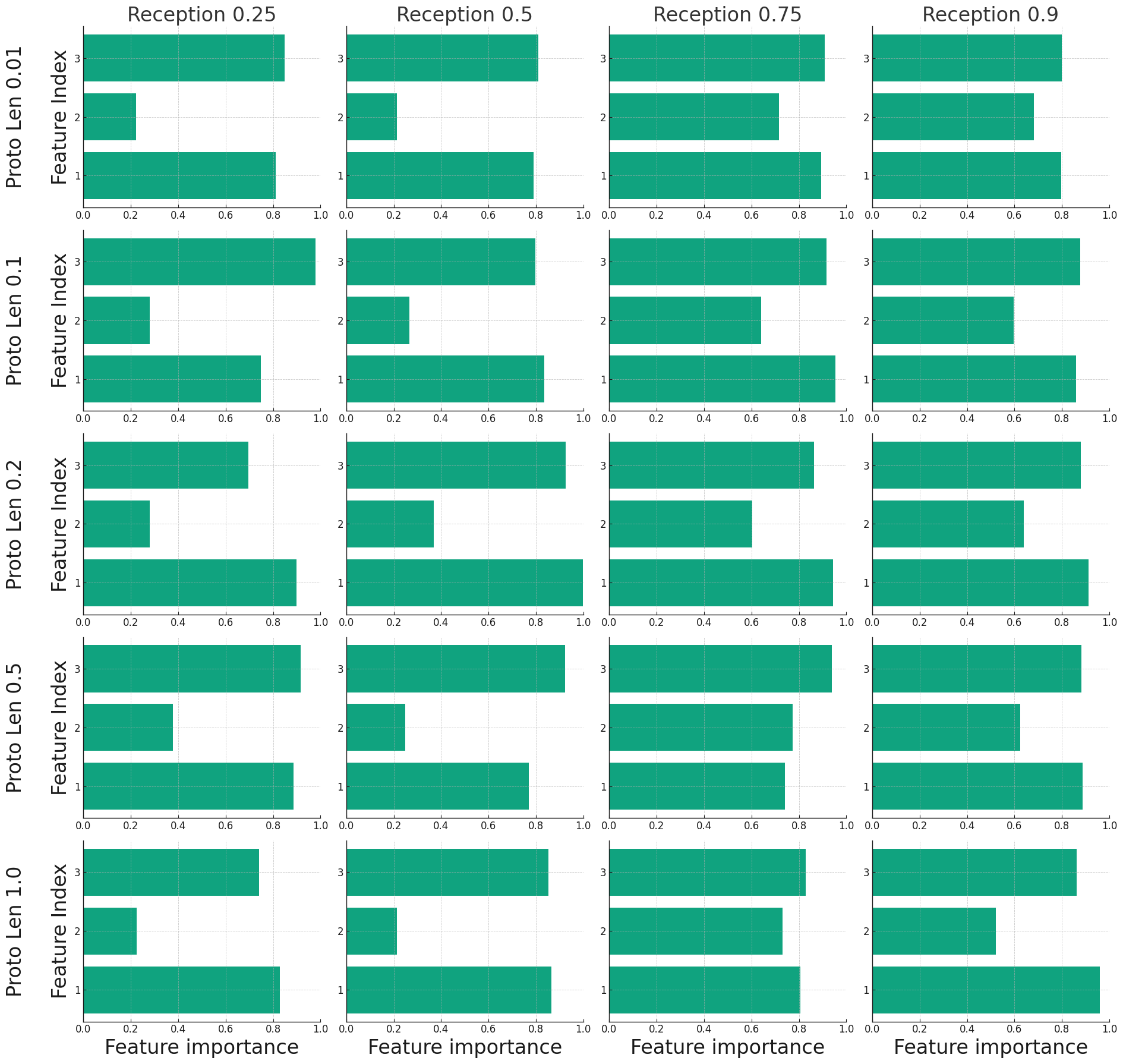}
        \end{minipage}
    }

    \caption{(contd)}
\end{figure}
\begin{figure}[t]\ContinuedFloat
    \centering

    LSST
    \subfloat{
        \centering
        \begin{minipage}[c][\height][c]{0.38\textwidth}
            \includegraphics[width=\textwidth]{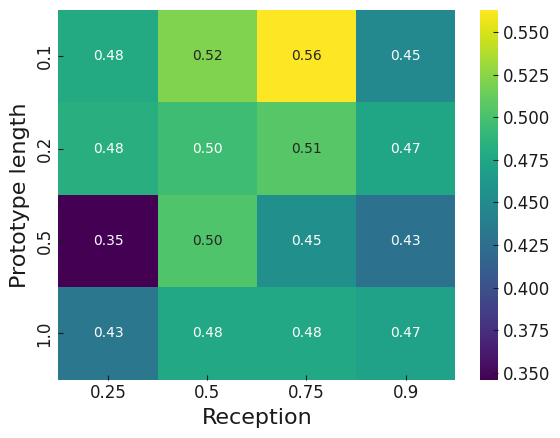}
        \end{minipage}
    }
    \subfloat{
        \centering
        \begin{minipage}[c][\height][c]{0.58\textwidth}
            \includegraphics[width=\textwidth]{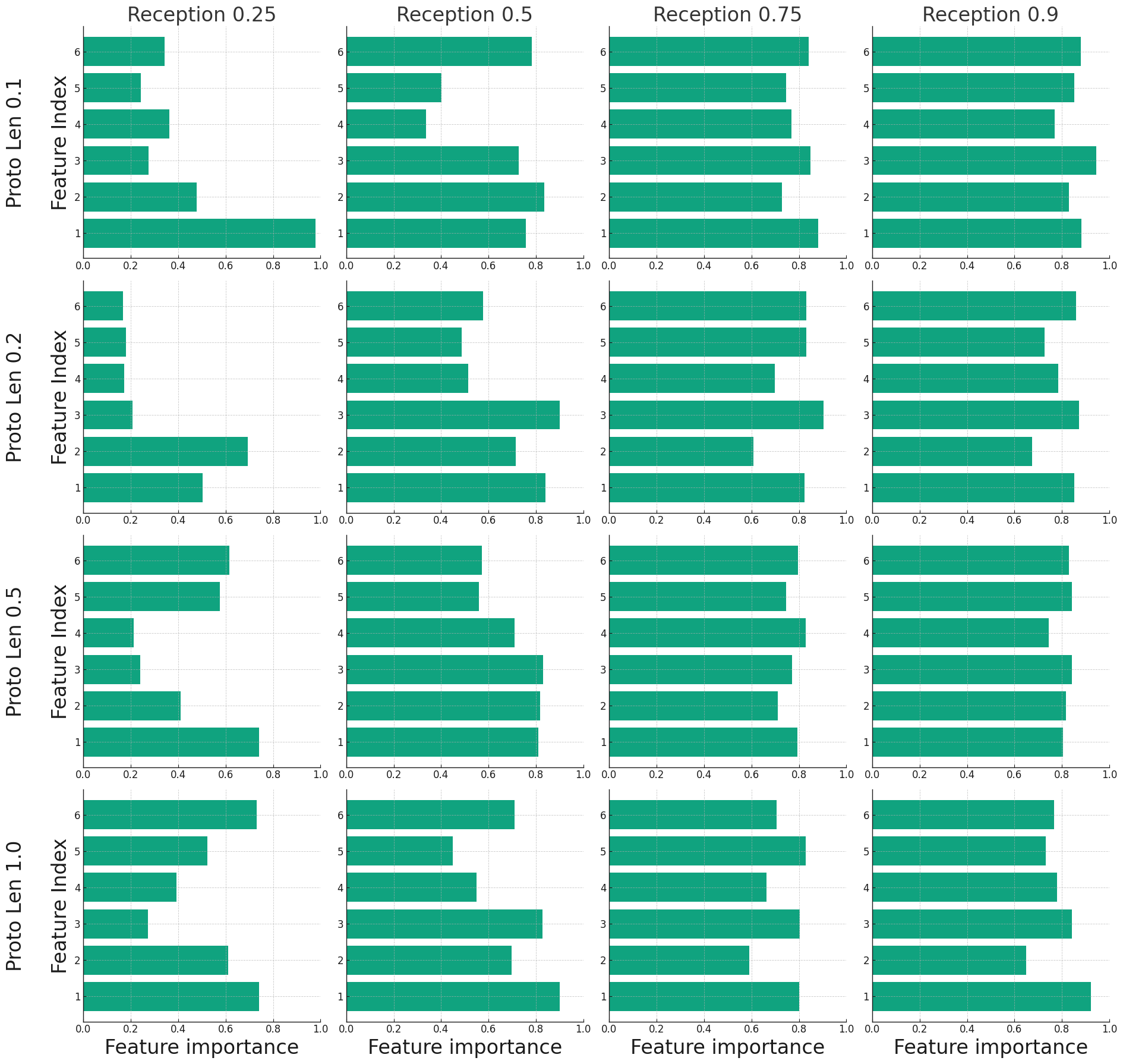}
        \end{minipage}
    }

    \bigskip
    SelfRegulationSCP1
    \subfloat{
        \centering
        \begin{minipage}[c][\height][c]{0.38\textwidth}
            \includegraphics[width=\textwidth]{assets/ablation/heatmaps/SelfRegulationSCP1.png}
        \end{minipage}
    }
    \subfloat{
        \centering
        \begin{minipage}[c][\height][c]{0.58\textwidth}
            \includegraphics[width=\textwidth]{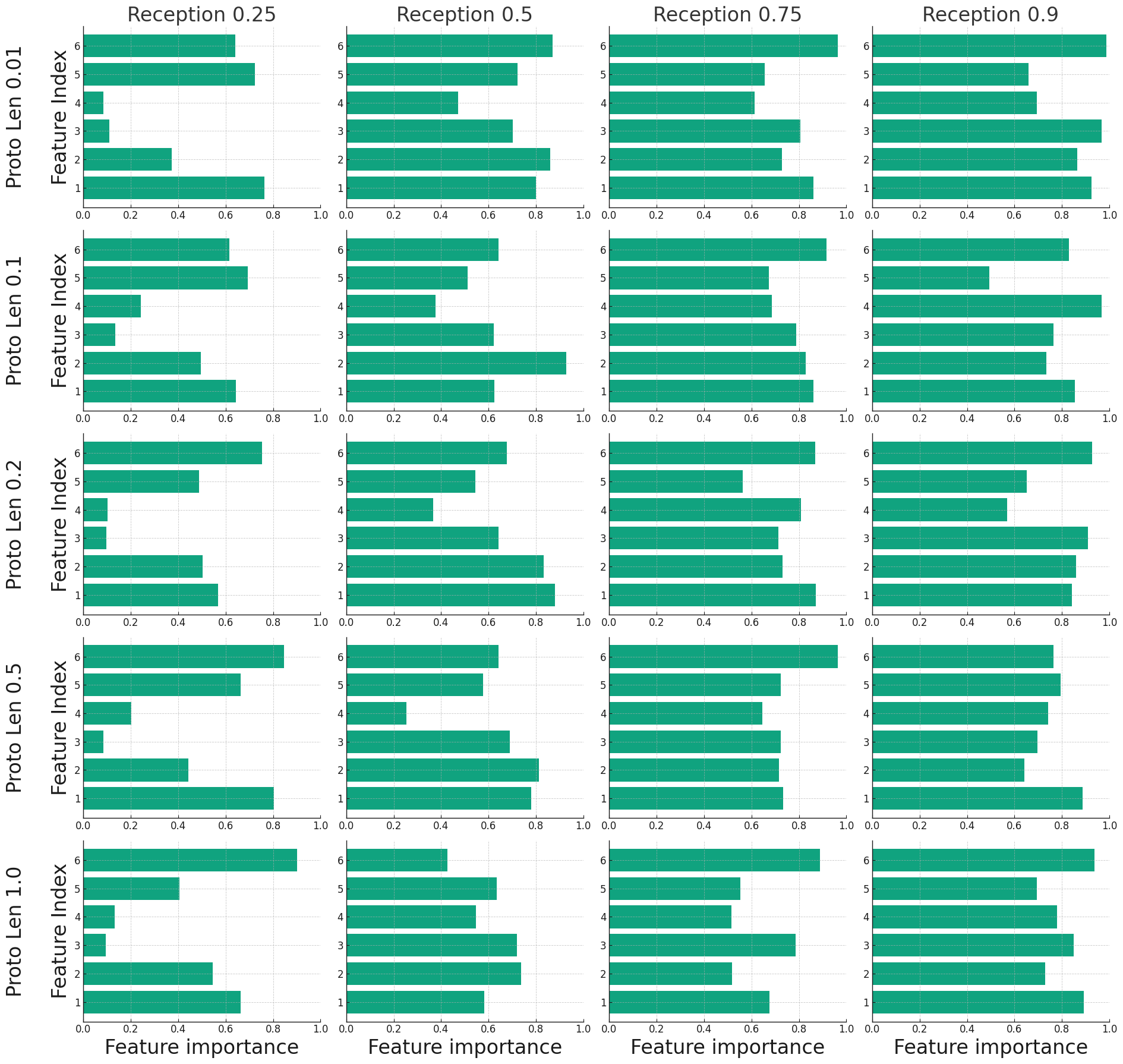}
        \end{minipage}
    }

    \caption{(contd)}
\end{figure}

\FloatBarrier

\subsection{Libras experiment prototypes}

\begin{figure}[hb]
    \centering
    \includegraphics[width=0.8\textwidth]{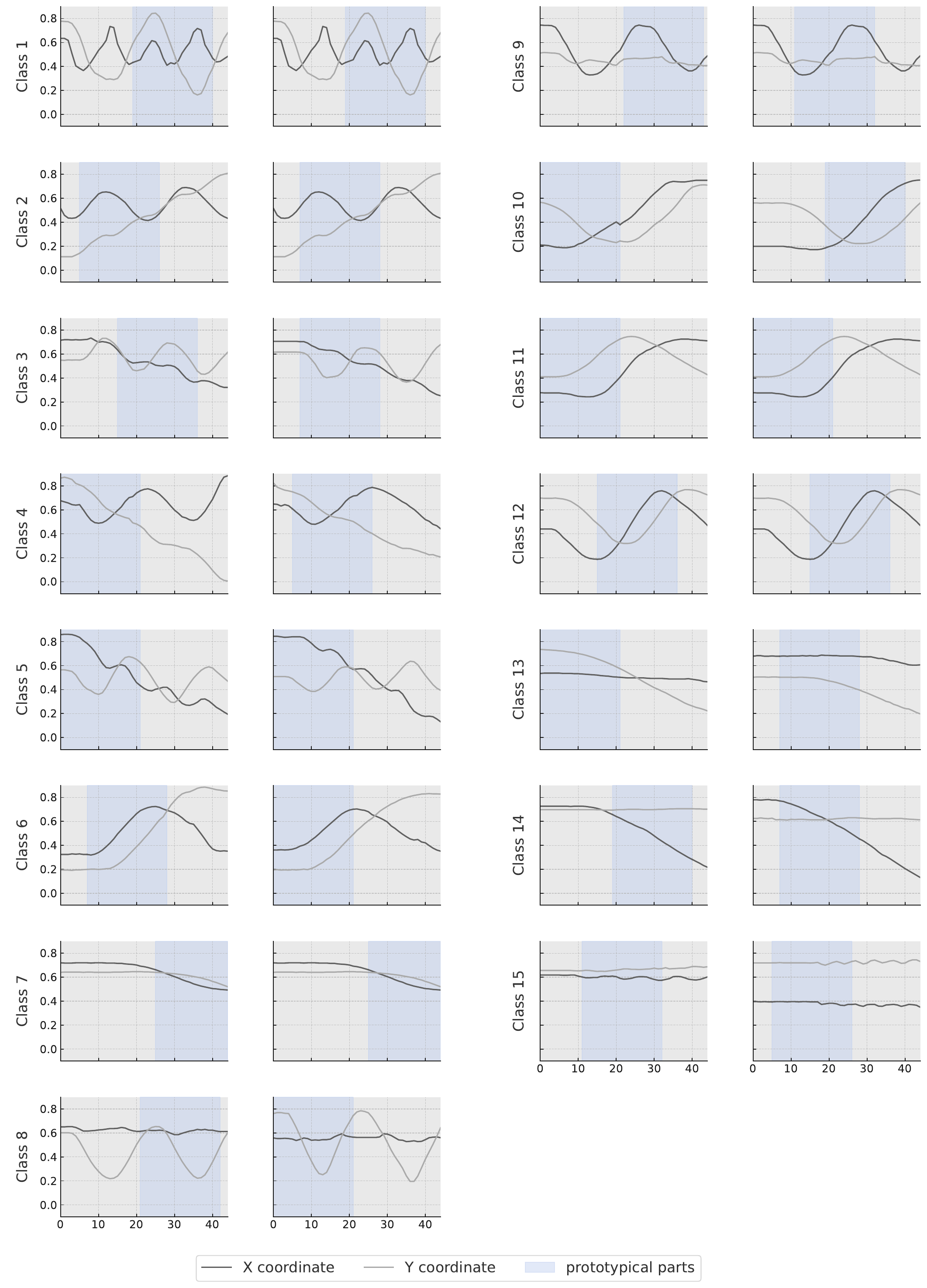}
    \caption{All prototypes learned on the Libras dataset shown as time-series (two channels per plot) with prototypical subsequences highlighted in blue. Each class had two prototypes, the prototypes were learned as a part of the experiment described in the main paper, $\lambda_{\text{last}}=3\times10^{-3}$, panels are grouped by class (15 classes).}
    \label{fig:libras-all-prototypes-1d}
\end{figure}

\begin{figure}[hb]
    \centering
    \includegraphics[width=\textwidth]{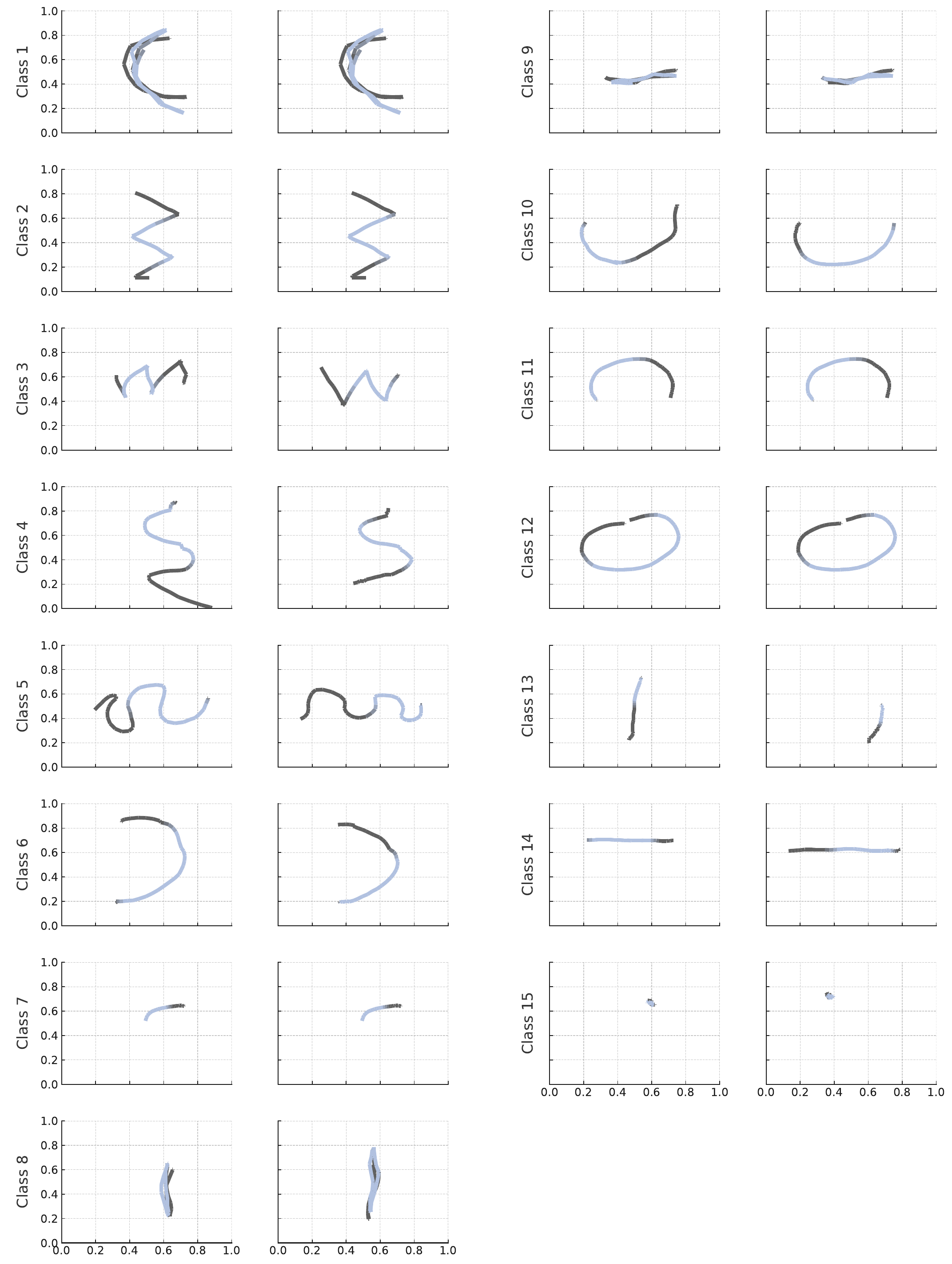}
    \caption{All Libras prototypes rendered as 2D hand trajectories with the prototypical segments highlighted in blue. Each class had two prototypes, the prototypes were learned as a part of the experiment described in the main paper, $\lambda_{\text{last}}=3\times10^{-3}$, panels are grouped by class (15 classes).}
    \label{fig:libras-all-prototypes-2d}
\end{figure}

\begin{figure}[t]
    \centering
    \includegraphics[width=0.98\textwidth]{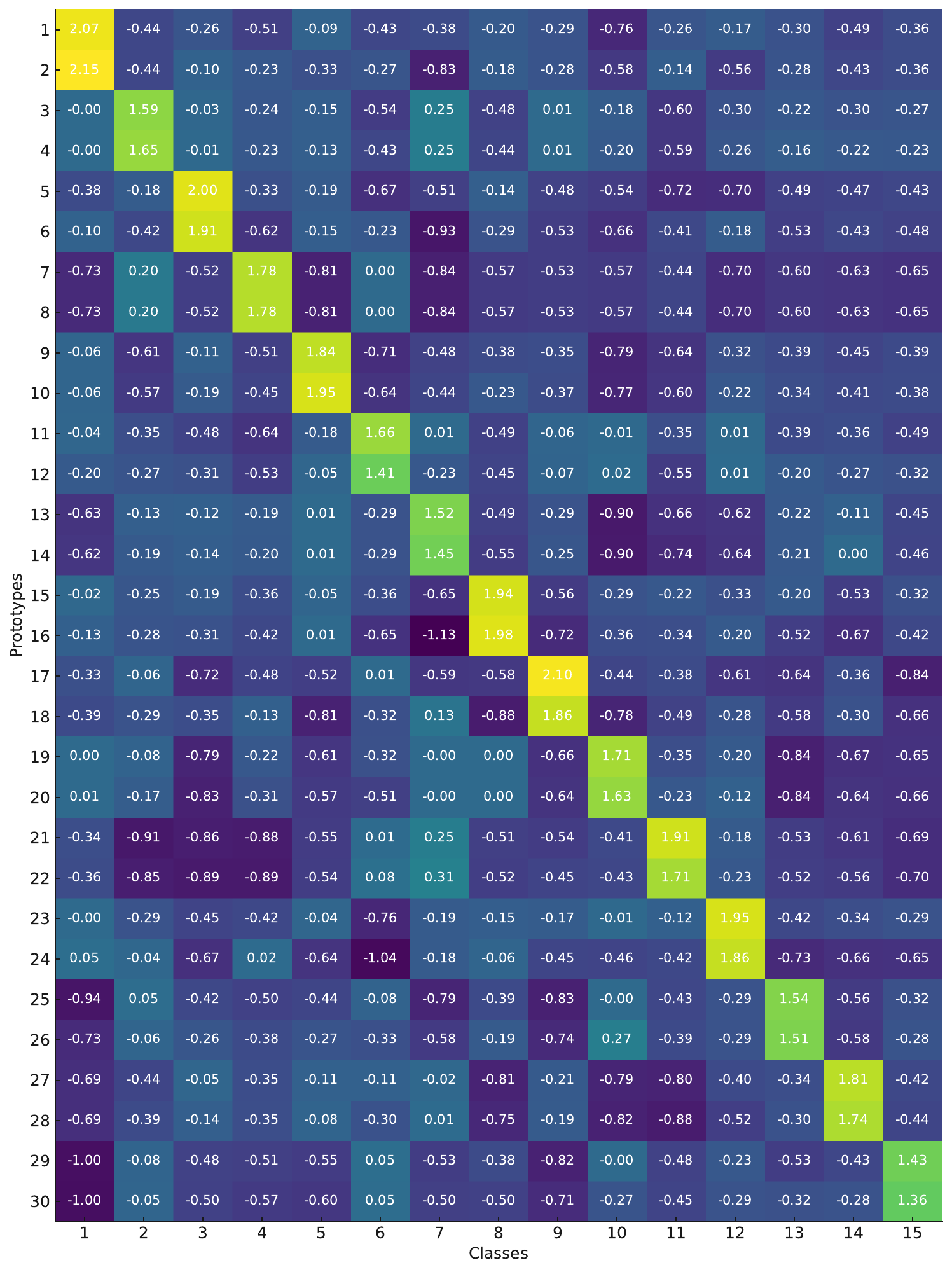}
    \caption{Last-layer weight matrix for the Libras experiment with two prototypes per class ($\lambda_{\text{last}}=1\times10^{-3}$).
    Columns correspond to classes (1-15) and rows to prototypes (1-30, ordered in class-wise pairs).
    Positive values (yellow) indicate "this resembles that" evidence; negative values (purple) indicate "does not resemble that".
    Biases are disabled in this layer. Compared with the main-text setting ($\lambda_{\text{last}}=3\times10^{-3}$, Fig.~\ref{fig:libras-ll-weights}),
    at this lower $\lambda_{\text{last}}$ setting, negative off-diagonal connections are widespread (many non-zero entries), indicating frequent use of supporting decisions with other-class prototypes.}
    \label{fig:libras-ll-weights-low}
\end{figure}

\begin{figure}[t]
    \centering
    \includegraphics[width=0.98\textwidth]{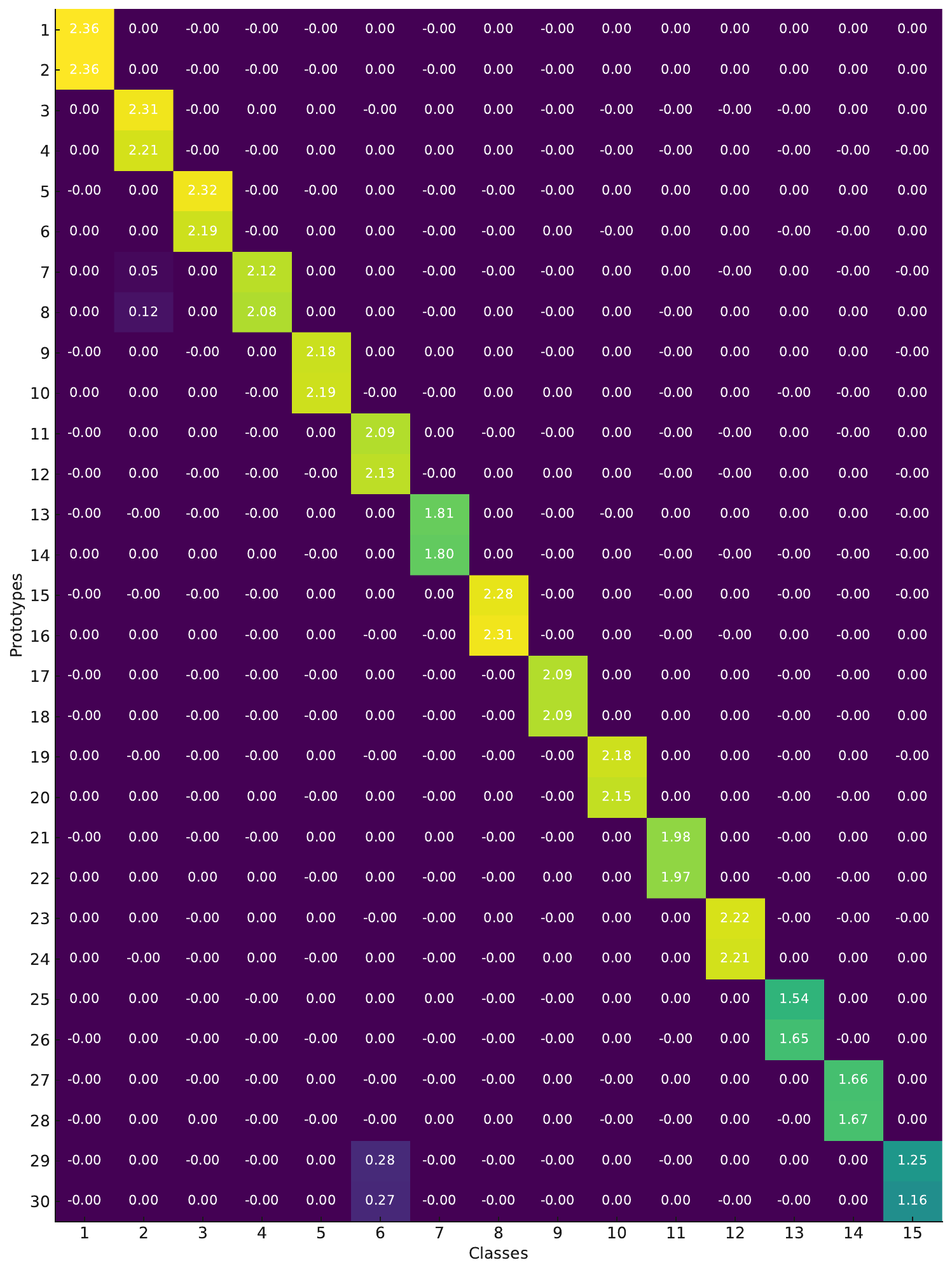}
    \caption{Last-layer weight matrix for the Libras experiment with two prototypes per class ($\lambda_{\text{last}}=1\times10^{-2}$).
    Columns correspond to classes (1-15) and rows to prototypes (1-30, ordered in class-wise pairs).
    Positive values (yellow) indicate "this resembles that" evidence; negative values (purple) indicate "does not resemble that".
    Biases are disabled in this layer.
    Relative to the main-text setting ($\lambda_{\text{last}}=3\times10^{-3}$; Fig.~\ref{fig:libras-ll-weights}), the higher $\lambda_{\text{last}}$ suppresses off-diagonal negatives almost entirely, yielding an almost purely diagonal, positive-evidence mapping.}
    \label{fig:libras-ll-weights-high}
\end{figure}

\end{appendices}

\end{document}